\newcommand{\CLASSINPUTinnersidemargin}{25mm}
\documentclass[12pt,journal,onecolumn,a4paper]{IEEEtran}

\usepackage{times}
\usepackage{xcolor}
\usepackage{soul}
\usepackage{xspace}
\usepackage{subfig}

\usepackage[colorlinks,pagebackref,linktocpage,draft,breaklinks]{hyperref}
\usepackage{url}
\usepackage{graphicx}

\usepackage{pgf}  
\usepackage{xfrac}
\usepackage{floatrow}
\usepackage{subfig}
\usepackage{enumitem}
\setlist{nosep} 

\usepackage{amsopn}

\usepackage{scrextend}

\newcommand{\comment}[1]{}

\usepackage[noadjust]{cite}
\renewcommand{\IEEEbibitemsep}{3pt plus 1.5pt}
\makeatletter
\IEEEtriggercmd{\reset@font\normalfont\fontsize{9.9pt}{10.40pt}\selectfont}
\makeatother
\IEEEtriggeratref{1}

\usepackage{color}
\definecolor{darkred}{rgb}{0.55, 0.0, 0.0}
\definecolor{orange}{RGB}{255,127,0}
\definecolor{brown}{RGB}{150,70,0}
\definecolor{green}{RGB}{127,255,127}
\definecolor{darkgreen}{RGB}{0,127,0}
\definecolor{blue}{RGB}{127,127,255}
\definecolor{lightblue}{RGB}{150,150,255}
\definecolor{darkblue}{RGB}{0,0,127}
\definecolor{red}{RGB}{255,90,90}
\definecolor{grey}{RGB}{127,127,127}
\definecolor{pink}{RGB}{255,180,180}

\newcommand{\red}[1][red]{\textcolor{red}{{#1}}}

\newcommand{\green}[1][green]{\textcolor{green}{{#1}}}
\newcommand{\grey}[1][grey]{\textcolor{grey}{{#1}}}

\usepackage{xcolor}
\usepackage{nicefrac}

\newcommand{\inserted}[1]{\textcolor{darkblue}{{#1}}}
\newcommand{\modified}[1]{\textcolor{darkred}{{#1}}}

\newcommand{\xaxis}{$x$-axis\xspace}

\newcommand{\Response}{\ensuremath{\mathit{R}}\xspace}

\newcommand{\gendiff}{\ensuremath{\mathit{\gamma}}\xspace}

\begin{document}
%
\title{Analysing Results from AI Benchmarks: \\ Key Indicators and How to Obtain Them\thanks{This report is a preliminary version of a related paper with title "Dual Indicators to Analyse AI Benchmarks: Difficulty, Discrimination, Ability and Generality", accepted for publication at IEEE
  Transactions on Games (DOI: \href{https://doi.org/10.1109/TG.2018.2883773}{10.1109/TG.2018.2883773}). Please refer to and cite the journal paper.}}

%

\author{Fernando~Mart\'inez-Plumed
        and~Jos\'e~Hern\'andez-Orallo\\
        {\texttt{\{fmartinez, jorallo\}@dsic.upv.es}}\\
        Universitat Polit\`ecnica de Val\`encia, Spain

}

%
%

\markboth{}%
{Mart\'inez-Plumed \MakeLowercase{\textit{et al.}}: Dual Indicators to Analyse AI Benchmarks: \\Difficulty, Discrimination, Ability and Generality}
%



\maketitle

\begin{abstract}
Item response theory (IRT) can be applied to the analysis of the evaluation of results from AI benchmarks. 
The two-parameter IRT model provides two indicators ({\em difficulty} and {\em discrimination}) on the side of the item (or AI problem) while only one indicator ({\em ability}) on the side of the respondent (or AI agent). In this paper we analyse how to make this set of indicators dual, by adding a fourth indicator, {\em generality}, on the side of the respondent. Generality is meant to be dual to discrimination, and it is based on difficulty. Namely, generality is defined as a new metric that evaluates whether an agent is consistently good at easy problems and bad at difficult ones. With the addition of generality, we see that this set of four key indicators can give us more insight on the results of AI benchmarks. In particular, we explore two popular benchmarks in AI, the Arcade Learning Environment (Atari 2600 games) and the General Video Game AI competition. We provide some guidelines to estimate and interpret these indicators for other AI benchmarks and competitions. 
\end{abstract}


%
\IEEEpeerreviewmaketitle

\section{Introduction}
%
%
%
%

The evaluation of AI systems has traditionally been done with one system evaluated on one single problem. Some of the early breakthroughs on draughts (checkers) in the 1950s \cite{Samuel59}, chess in the 2000s with \emph{Deep Blue} against the human chess champion Garry Kasparov \cite{deepblue2002}, or even more recent ones, such as the 2010s IBM's program Watson winning the 
{\em Jeopardy!} TV quiz \cite{ferrucci2010building,ferrucci2013watson}, or the 
game of Go \cite{AlphaGo16}, were considered on the sole success of a very specialised task.

However, as the success of one system for one task cannot be extrapolated for other tasks, there is an increasing interest in the evaluation on a 
set of problems or applications. In order to prevent systems from specialising to these benchmarks, these try to include as many problems and as diverse as possible. Also, because of the maturity of some subfields in AI, many different techniques are available. This leads to a situation where many techniques are usually confronted with many problems. This is what we usually find 
in the experimental section of many technical papers and, especially, on running competitions. 
Examples are the UCI machine learning repository \cite{Lichman:2013}, the ICAPS planning and scheduling competitions \cite{vallati20152014}
 or the ImageNet challenges \cite{russakovsky2015imagenet} (see \cite{hernandez2016aire} for a more complete list of benchmarks and competitions). 
%

As the result of a virtuous circle with these new benchmarks, AI is able to generate much more general-purpose, adaptive and successful behaviours,  primarily in (video) games. For instance, deep reinforcement learning and other approaches are now able to perform extremely well in board games (e.g., \cite{AlphaGo16}) and relatively well in many arcade games (e.g., \cite{mnih2015human}). This brings the potential to use them as \textit{non-player characters} (NPC) or non-human opponents for more complex games in the future \cite{yannakakis2015panorama}, where the same architecture can be retrained for different games, without the effort of designing specific NPCs for each game. However, in these more generic scenarios, we do not know how to analyse their behaviour beyond specific performance, especially when we want to compare different approaches for a range of games. In particular, apart from specialised agents, it would be useful to have generic algorithms that can produce relatively good NPCs for games that look of easy or medium difficulty, instead of  those that are very good, or specialised, at some hard games but very poor at many easy others.

AI is also now paying attention to systems that solve several tasks at a time \cite{hernandez2016aire,orallo2017}. Indeed, a popular setting for general-purpose evaluation today is a collection of games under an interactive scenario, where agents can perceive and act, and are rewarded when they make good choices. Many different platforms have recently appeared in this regard \cite{hernandezCosmos2017}, laying special emphasis on the use of 2D/3D videogames for AI evaluation \cite{hernandez2010measuring} and attracting mainstream attention \cite{castelvecchi2016tech}.

Two representative examples are the \textit{Arcade Learning Environment} \cite{bellemare2015arcade}, a collection of Atari 2600 games; and the \textit{General Video Game AI} (GVGAI) competition \cite{perez20162014}, a benchmark that comprises a large number of real-time 2D grid games such as puzzles, shooters and classic arcades. Both ALE and GVGAI are remarkable benchmarks that allow us to observe the performance of AI agents on a multiplicity of problems. They have both received plenty of interest and have become a reference for AI experimentation and evaluation in the past few years.
The popularity of these AI benchmarks have also produced a good number of results that can now be analysed in hindsight and used to better understand not only these benchmarks, but also general-purpose AI overall. This analysis can be understood from the viewpoint of the AI systems (and how to improve AI techniques) but also from the viewpoint of the problems (and how to improve the benchmarks). In this paper we argue that using performance is insufficient to get a proper insight of what the systems are able to do (and how they achieve it) and what the problems in the benchmark are evaluating. 

In this paper we present two pairs of key indicators that can help us understand the results in AI benchmarks in a more informative way. On the one hand, we claim difficulty and discrimination as key indicators for AI problems. On the other hand, we postulate ability and generality as key indicators for AI systems. Namely,

\begin{itemize}
\item Looking at the problems, inferring a difficulty indicator helps us control whether we are evaluating a proper range of difficulties, and clarifies that we expect a general systems to perform well for almost all easy problems before we can direct our progress towards areas of higher difficulty. Relatedly, it is also important to detect whether difficult problems are only solved by able systems, and not by chance or specialisation by very poor systems. This is the notion of discrimination, which will spot that some problems may be useless, or even detrimental, for an efficient and robust evaluation. Taking into account the increasing computing demands of training and evaluation for recent algorithms, any understanding of what the key tasks are can imply an important contribution for AI researchers. 

\item
Looking at the systems, ability gives us a different perspective from performance, as it considers the difficulty of the problems, instead of a simple average. But the most novel insight comes from seeing whether a system --motivated by increasing performance-- focuses on a big pocket of problems while neglecting some other smaller pockets. It is of key importance, however, and widely overlooked in AI, that {\em we must understand generality in the context of difficulty} \cite{orallo2017}. In other words, if a system covers some of the low-hanging fruits but excludes others, we may suspect there is some specialisation. On the contrary, if a system covers all low-hanging fruits and almost none of the hard problems, we can usually infer some kind of systematic generality in the behaviour of the system. Ultimately, it is crucial for AI researchers to know whether they are progressing through generality or through the exploitation of specific subfamilies of problems.
\end{itemize}

\noindent The analysis under these key indicators represents a novel way of understanding not only benchmark results in AI \cite{hernandezCosmos2017}, but also video game competitions (e.g., \textit{Super Mario Bros} \cite{togelius2013mario}, \textit{Angry Birds} \cite{renz2015aibirds} or \textit{StarCraft} AI competitions) as well as the existing architectures for multi-purpose game agents and bots addressing them \cite{hosu2015comparative, khalifa2017multi}
. This kind of assessment may have a huge impact on how players and competitions are designed and how the results of the AI systems (and humans) are interpreted. We are not claiming that these two pairs 
of indicators are necessarily giving us the most complete information (this is ultimately given by the whole data), but they are a good trade-off between monolithic indicators (and limited insight) and too many indicators (and strong overlap).

In this paper, we obtain these indicators in different ways. Some of them (difficulty, discrimination and ability) are estimated through simple models inherited from Item Response Theory (IRT), a powerful technique from psychometrics \cite{embretson2000item}. Generality, as newly defined in this paper, is derived from the dispersion statistics, but taking difficulty into account. In all cases --and this is important to note--, the indicators are populational, i.e., they depend on a set of AI systems and a set of AI problems. Consequently, when we change the population, the obtained indicators may change as well.

The rest of the paper is organised as follows. Section \ref{measures} presents the indicators we propose for analysing the results of AI benchmarks. Section \ref{Data} describes the result data we will explore: the ALE and GVGAI problems and the AI techniques used to solve them. The estimation of the indicators for these benchmarks and how they can be used to understand the behaviour of problems and systems, is seen in sections \ref{TaskAnalysis} and \ref{TechAnalysis} respectively. Section \ref{Discussion} summarises the main findings and contributions, describes some limitations and guidelines for evaluation, and discusses the future work.

\section{Key indicators for AI benchmark results} \label{measures}

Whenever a benchmark is built, its creators usually consider a set of problems or tasks that are representative of the kinds of applications we want to progress on. One way of looking at this progress is in terms of overall performance. If the set of systems or agents $j$ is $\Pi$, with $|\Pi|=m$, and the set of problems, tasks or items $i$ is $M$, with $|M|=n$, we can have a measure of the result (the {\em response}) of each system $j$ for each problem $i$, as $\Response_{j,i}$, making up an $m\times n$ result matrix\footnote{We follow the usual convention in IRT with items being columns referred to by index $i$.}. Then, for a single system $j$, we can calculate its (weighted) average performance $\sum_j w_i \Response_{j,i}$, where $w_i$ are weights given to problems. The first thing this assumes is that performances are commensurate, with the weights giving more or less relevance to some problems depending on their importance. In this case, we would say that a system $A$ is better than a system $B$ if the (weighted) average performance is higher. In some other cases, if the performances are not commensurate, one can be satisfied with a binary comparison by how many wins/ties/losses there are in their performances. If we want to compare more than two systems at the same time, we would use rankings instead, which can be produced from the aggregated performance, from a count of pairwise comparisons or in other ways. There are many other variations, especially when we want to apply some statistical tests on the results, but all of them are based on some notion of aggregated or comparative performance.

However, there is another way of looking at this. One can consider that not all problems have the same difficulty. That does not mean that difficult problems should count more than easy problems, but that difficulty should be taken into account in any notion of quality of a system, and most importantly, in any measure of progress. 
Indeed, a system behaving well on difficult items but poorly on easy items would certainly be a  strange specimen. In a way, we not only expect a positive monotonicity between the quality of a system and the probability of a correct response, but also some kind of positive monotonicity between the quality of a system and the difficulty of the problems it can likely solve. This observation suggests a completely different way of analysing results, which led to IRT, as we introduce next.


\subsection{IRT in AI: ability, difficulty and discrimination} \label{IRT}

Item response theory (IRT) \cite{embretson2000item} has mainly been used in educational testing and psychometric evaluation in which examinees' ability is measured using a test with several questions (i.e., items). In essence, IRT is a set of mathematical models that describe the relationship between a latent trait of interest and the respondents' answers to individual items, where the probability of a response for an item is a function of the examinee's ability and some item's parameters. There are models developed in IRT for different kinds of response, but we will focus on the dichotomous models where responses can be either correct or incorrect. Multiple choice items (more than two options) can also be considered dichotomous since they can still be scored as correct/incorrect.

In this context, let $\Response_{j,i}$ be the binary result of a respondent $j$ to item $i$, with $\Response_{j,i}=1$ for a correct response and $\Response_{j,i}=0$ otherwise. Let $\theta_j$ be the ability or proficiency of $j$, and let us imagine for a moment that this value is known. Now, assuming that the result only depends on the ability of the respondent, and we assume a particular value for respondent $j$, we can express the result as a function of $i$ alone, i.e. $\Response_{i}$. For the basic 3-parameter (3PL) IRT model, the probability of a correct response on an item given the examinee's ability is modelled as a logistic function:

\begin{equation}\label{eq:3pl}
P(\Response_{i}=1|\theta_j)= c_i + \frac{1-c_i}{1+e^{-a_i(\theta_j-b_i)}}
\end{equation}

\begin{figure}[!h]
\begin{center}
\includegraphics[width=0.50\columnwidth]{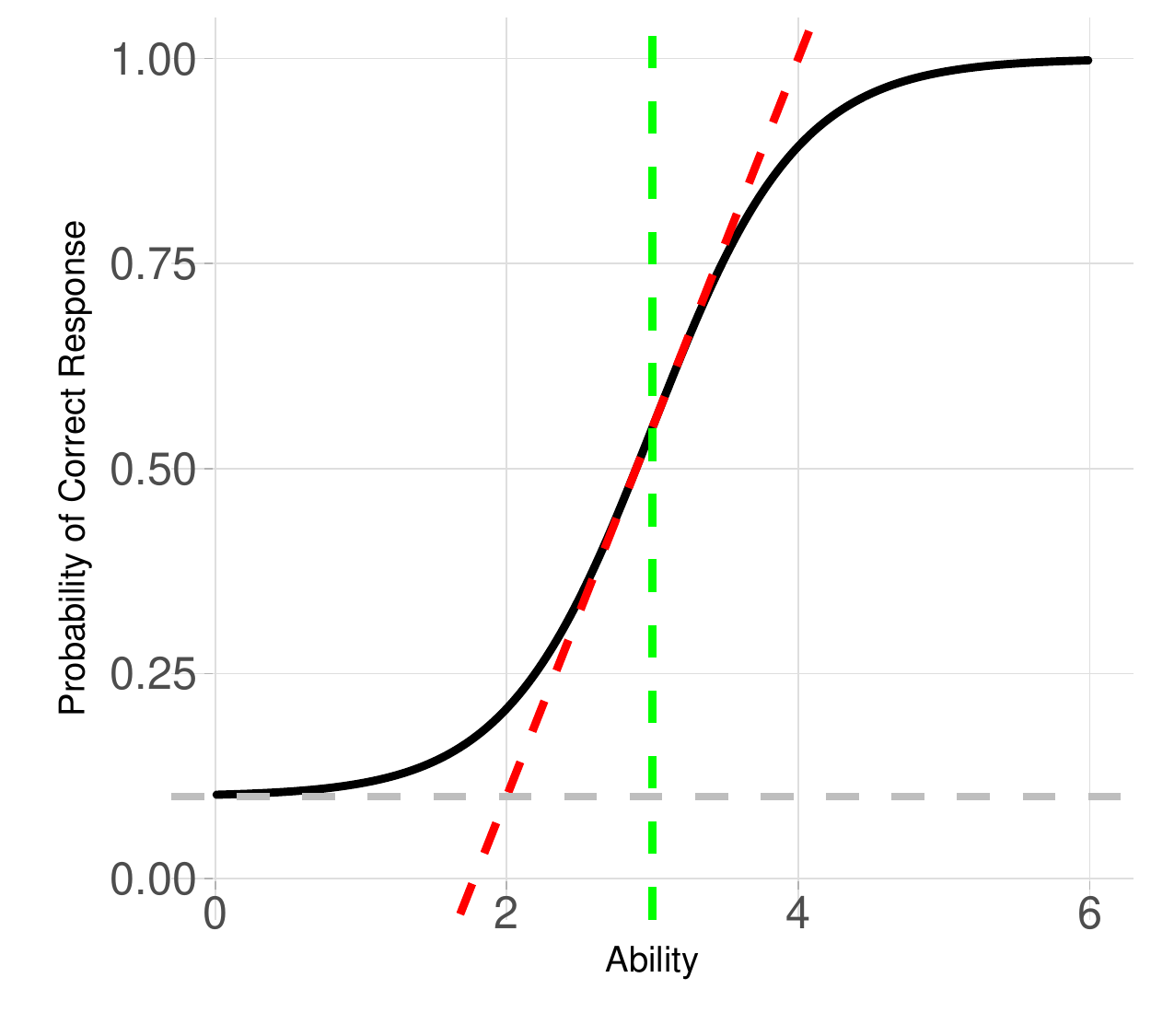}
\caption{Example of a 3PL IRT ICC curve
, with slope $\red[a]= 2$%
, location parameter $\green[b]= 3$ 
and guessing parameter $\grey[c]= 0.1$. 
}\label{fig:3PL_example_A}
\end{center}
\end{figure}

\noindent The above model provides an {\it Item Characteristic Curve (ICC)} (see Fig.~\ref{fig:3PL_example_A}) with three parameters:

\begin{itemize}
\item \textit{Difficulty} ($b_i$): it is the location parameter of the logistic function and can be seen as a measure of item difficulty. When $c_i=0$, then $P(\Response_{i}=1|b_i)=0.5$. 

\item \textit{Discrimination} ($a_i$): it indicates the steepness of the function at the
location point. For a high value, a small change in ability can result in a big change in the item response. Alternatively we can use the slope at location point, computed as $a_i(1-c_i)/4$ to measure the discrimination value of the instance.

\item \textit{Guessing} ($c_i$): it represents the probability of a correct response
by a respondent with very low ability ($P(\Response_{i}=1|-\infty)=c_i$).This is usually associated to a result given by chance.
\end{itemize}

\noindent The basic IRT models can be simplified to two parameters (e.g., assuming that $c_i = 0$), or just one parameter (assuming $c_i = 0$ and a fixed value of $a_i$, e.g., $a_i = 1$).

In our adaptation of IRT, an item in IRT can be identified with a problem or task in AI (e.g., an ALE or GVGAI game), and an individual, subject or respondent can be identified with an AI method, technique or system \cite{martinez2017egpai}. 
While a guessing parameter might be meaningful in some AI problems (e.g., classification \cite{IRT:lmce2015,martinez2016making,martinez2019IRT}), it is not appropriate when a random agent is expected to score poorly (e.g., in a videogame). On the contrary, the discrimination parameter is very informative about whether a particular instance is aligned with ability (i.e., to detect a negative monotonicity between the quality of a system and the probability of a correct response).

For Item Response Theory, the difficulty and discrimination of the items are considered latent traits that may be unknown, and they have to be estimated from the result matrix. Similarly,
the ability of an individual is considered a latent trait that can also be estimated based on her responses to discriminating items with different levels of difficulty. Respondents who tend to correctly answer the most difficult items will be assigned to high values of ability. Difficult items in turn are those correctly answered only by the most proficient respondents. Notice that ability and difficulty appear subtracted in the exponent of the logistic model in Eq.~\ref{eq:3pl}, so they are on the same scale, which gives these two parameters a dual character (e.g., an agent of ability 4 has 0.5 probability of being correct for an item of difficulty 4). Also many models assume that both parameters follow a normal distribution.

Straightforward methods based on maximum-likelihood estimation (MLE) can be used to estimate either the item parameters (when respondent abilities are known) or the abilities (when item parameters are known). A more difficult, but common, situation is the estimation when both the item parameters and respondent abilities are unknown. In this situation, an iterative two-step procedure, the Birnbaum's method \cite{Birnbaum:book68}, can be adopted for dichotomous items:

 \begin{itemize}
\item (1) Start with initial values for abilities $\theta_j$ (e.g., random values or the number of correct responses). 
\item (2) Estimate the model parameters, assuming the abilities of the previous step.
\item (3) Estimate the abilities $\theta_j$, assuming the model parameters in the previous step.
\item (4) Until stop condition, go to (2).
\end{itemize}

\noindent Some implementations iterate the above process a fixed number of times (1,000 in {\ttfamily{ltm}} R package\footnote{\label{ltm}\url{https://cran.r-project.org/web/packages/ltm/}}) and/or stop when the model's goodness of fit reaches a given threshold. In Birnbaum's method, the fit of the model is based on approximate marginal maximum likelihood, using the Gauss-Hermite quadrature rule for the approximation of the required integrals. Item parameters and respondent abilities are alternatively estimated in this iterative process, and overall they are derived only based on a set of observed responses to items, with no previous knowledge about the true ability of the respondents.   

\comment{
\begin{figure}[!h]
\begin{center}
\includegraphics[width=0.60\columnwidth]{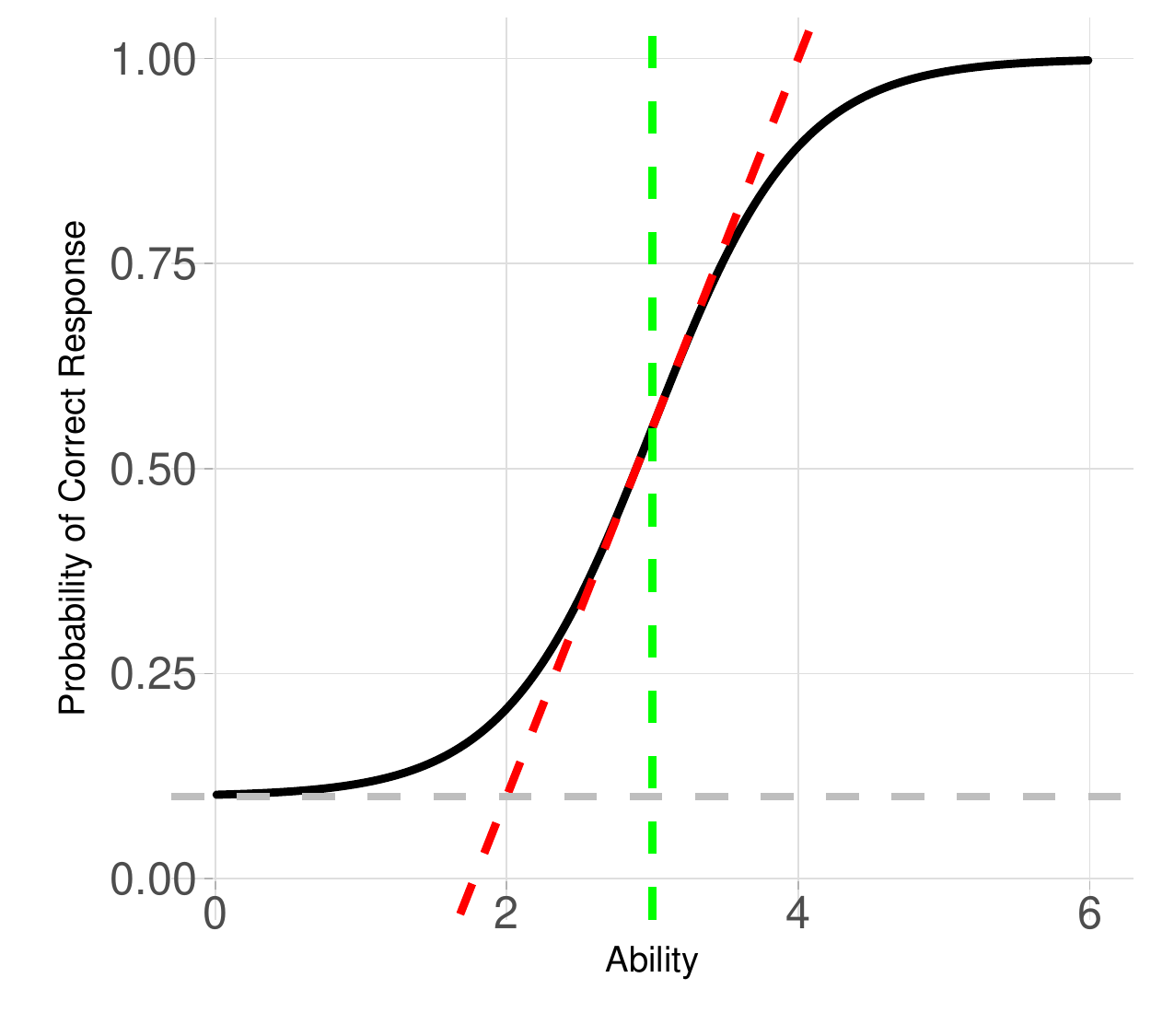}

    \vspace{0.4cm}
    
\includegraphics[width=0.60\columnwidth]{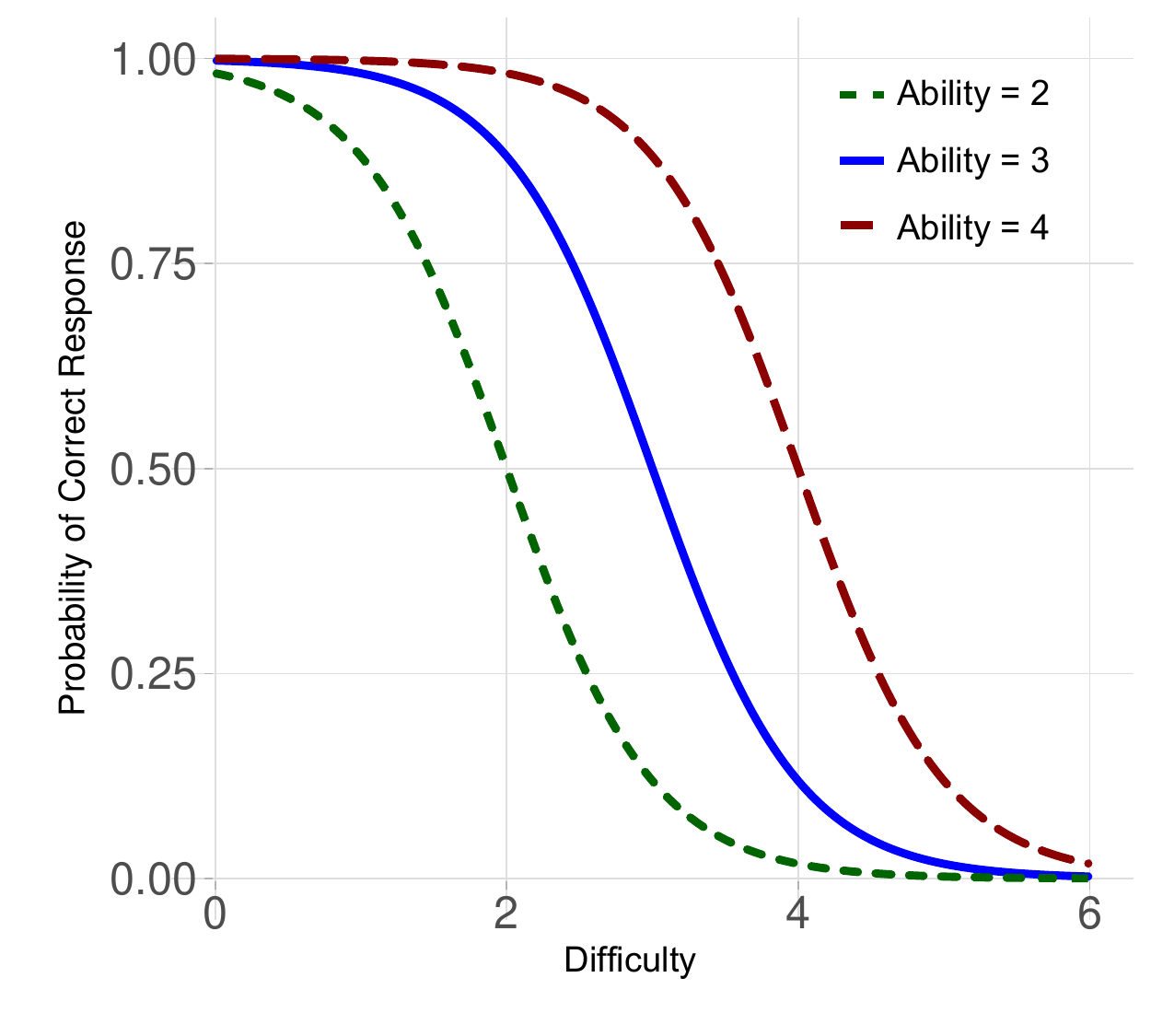}

\caption{(Top) Example of a 3PL IRT ICC curve
, with slope $\red[a]= 2$
, location parameter $\green[b]= 3$ 
and guessing parameter $\grey[c]= 0.1$ 
(Bottom) Example of PCC curves 
with different abilities. 
}\label{fig:3PL_example}
\end{center}
\vspace{-0.1cm}
\end{figure}
}

The key assumption, in any case, apart from the particular model family and the parameter scales and distributions, is monotonicity. Fig.~\ref{fig:3PL_example_A} shows an item characteristic curve, where the probability of correct response grows monotonically as a function of the ability of a classifier. 

A positive (i.e., increasing) monotonicity is captured by a positive discrimination parameter. We actually expect more able systems to perform better than less able systems for that item. If this is not the case, when discrimination is negative, we have an unusual problem (and abstruse item, in the IRT terminology). If the problems are well selected (or well filtered) we should not have negative discriminations. When comparing positive discriminations, higher positive values (steeper curves) indicate that the item is very informative around its level of difficulty (very discriminative in the region where a high slope takes probability from low to high values). On the contrary, low positive discrimination (flatter curves) means that the item is informative in a wider range of abilities, but less crisp near the difficulty parameter.

\begin{figure}[!h]
\begin{center}
\includegraphics[width=0.5\columnwidth]{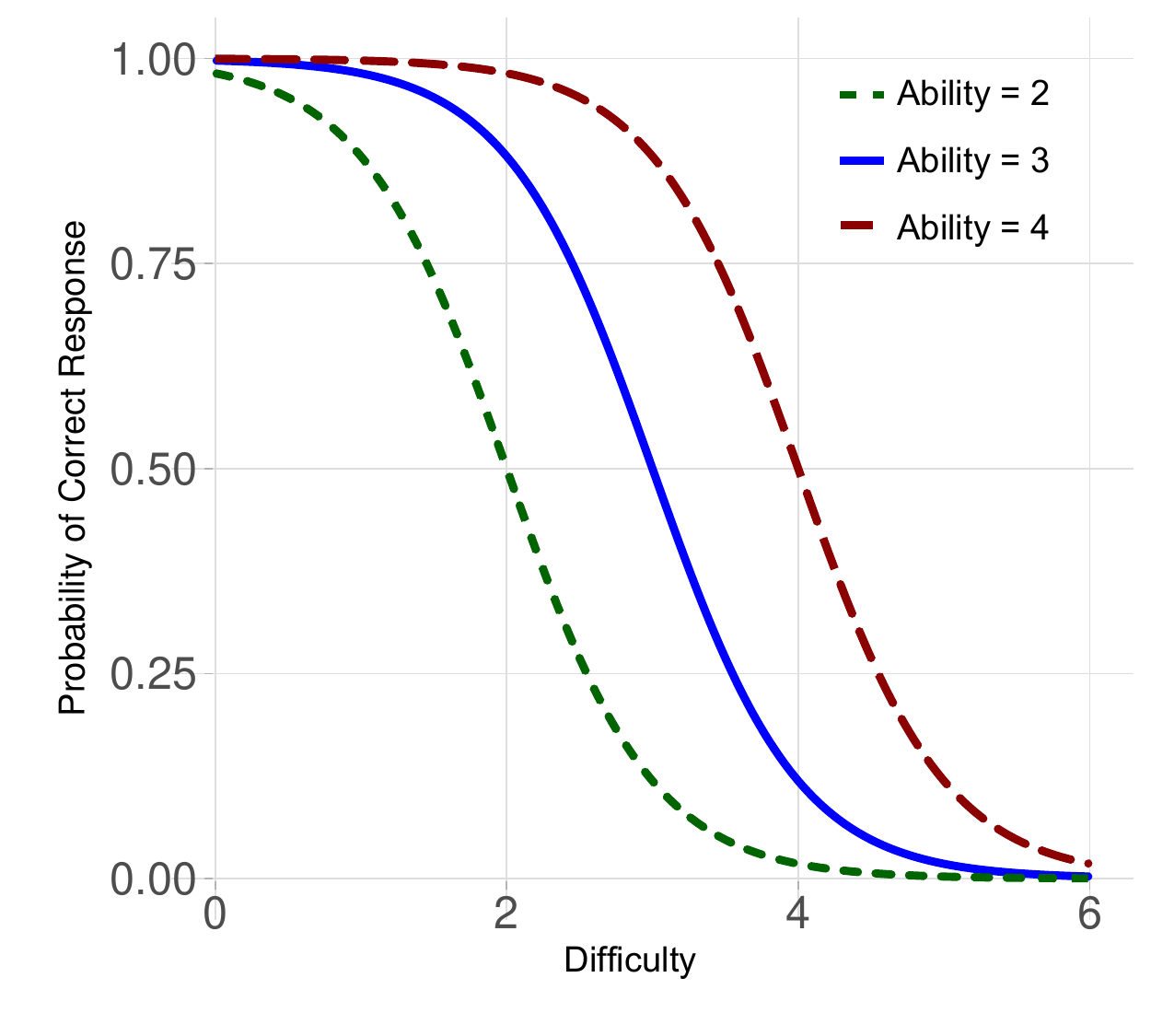}

\caption{Example of PCC curves with different abilities.}\label{fig:3PL_example_B}
\end{center}
\vspace{-0.1cm}
\end{figure}

In a dual way, for a given ability we can plot the probability of correct response against difficulty. Fig.~\ref{fig:3PL_example_B} shows three person characteristic curves (PCC) for three agents with different abilities. For AI we will call them {\em agent characteristic curves}. We expect these curves to be decreasingly monotonic, with very able systems being good at easy problems and decaying later than less able systems.

\subsection{Precluding specialisation in AI: generality} \label{Gen}

The dichotomy between narrow and general AI has caused a long and controversial debate we will not reproduce here. However, even when we restrict to particular domains, there is usually the very good argument that we cannot expect a general system to be good at all possible problems. For some classes of problems, it is possible to build a system that is valid for all the problems in the class, but the situation becomes trickier for some other classes. This is especially the case for very open classes of problems such as video games. The idea of thinking of a {\em general} system that excels at all of them seems infeasible for many areas and benchmarks. This means that systems will fail at some tasks.

The subset of problems a system fails at can be completely random or can follow some pattern. If this pattern represents some particular characteristics these problems do (not) have, we can infer that the system has somewhat specialised in favour of (or against) that pattern. But if the pattern is related to the resources or the difficulty of the problem, we tend to consider this as a general adjustment between possibilities and resources. For instance, a calculator that could only multiply even numbers is not considered very general, whereas an ordinary pocket calculator is still considered general for multiplication, even if it fails for all the large numbers that go beyond its digit capacity. This suggests that the crux of the question about generality is capacity, or {\em difficulty.} One {\em can} actually be good at all (or almost all) problems up to certain difficulty, even for very broad problem classes. 

This is related to some fundamental questions such as whether it is possible to build a system that excels at all possible problems. The no-free-lunch theorems \cite{wolpert1997no,wolpert2012no} argue that this is not possible, if one assumes block-uniformity for all possible problems. But this assumption will not hold if we order problems by some metric of difficulty (making very difficult problems less likely) and hence we look for agents that are good up to a certain level of difficulty. 
Actually, it is not a surprise that one finds free lunches, with systems that work generally well, when problems are built in such a way that resources (and difficulty) is not a completely unbounded and random variable, as happens with benchmarks in AI, or other games that were originally conceived for humans, which are resource-bounded systems. For instance, systems can be better than others overall, as observed by \cite{ashlock2017general} for an actual video game benchmark.  
As we will see, only when the notion of difficulty is introduced (as we have done with IRT in the previous subsection), the analysis of generality becomes really meaningful. 

To make the point even clearer, let us start with a notion that ignores difficulty, and we will introduce a version that does consider difficulty  afterwards. Ignoring difficulty or any other parameter of the problems, one can simply introduce a measure of dispersion. Let us denote by $\sigma_j^2$ the populational variance of results for system $j$:

\begin{equation}\label{eq:variance}
\sigma_j^2 =  \frac{\sum_i(\Response_{j,i} - \bar{\Response_j})^2}{n}
\end{equation}

\noindent where $\bar{\Response_j}$ is the average result for system $j$. 
Considering this variance as an extra, informative, parameter, we could simply define 
a measure of {\em regularity} as the inverse of the variance. In an AI benchmark, such as ALE or GVGAI, one system would be regular if their results have low variance. If a system has very good results on some problems but very bad results on others, even if the overall quality is good, the regularity would be low.

For binary responses, we have a Bernouilli distribution, which means that the variance is reduced to just $\bar{\Response_j} \cdot (1- \bar{\Response_j})$. Consequently, we would not need an extra parameter for the dispersion of results for an agent, as variance, and hence regularity, would just be a function of the average performance $\bar{\Response_j}$ of the agent. This is one reason why an extra fourth parameter  is not usually considered in the binary models in IRT. However, if the models are not binary, things are different.

\begin{table}[!ht]

\centering
\resizebox{0.5\columnwidth}{!}{%
\def\arraystretch{1.2}%
\begin{tabular}{llccc}
  
  \textbf{Model} 
  & $\mathbf{\bar{\Response_j}}$ & $\mathbf{\sigma_j^2}$ & $\sfrac{1}{\mathbf{\sigma_j^2}}$\\ 
  \hline
   $Constant_{[0]}$ & 
   0.00 & 0 & Inf \\ 
   $Constant_{[1]}$ &  
   1.00 & 0 & Inf \\ 
   $Constant_{[0.25]}$  &  
   0.25 & 0 & Inf \\ 
   $Constant_{[0.5]}$   &  
   0.50 & 0 & Inf \\ 
   $Constant_{[0.75]}$  & 
   0.75 & 0 & Inf \\ 
   $Categorical_{[0.3:0.5,0.4:0.5]}$ 	 & 
   0.35 &   0.00 & 400.00 \\ 
   $Categorical_{[0.7:0.5,0.8:0.5]}$ & 
   0.75 & 0.00 & 400.00 \\ 
   $Categorical_{[0.6:0.5,0.9:0.5]}$  & 
   0.75 & 0.02 & 44.44 \\ 
   $Categorical_{[0:0.3,1:0.7]}$ &  
   0.70 & 0.21 & 4.76 \\ 
   $Categorical_{[0.25:0.3,1:0.7]}$  &  
   0.78 & 0.12 & 8.47 \\ 
   $Categorical_{[0.5,1]}$ & 
   0.85 & 0.05 & 19.05 \\ 
   $Categorical_{[0.75:0.3,1:0.7]}$  & 
   0.92 & 0.01 & 76.19 \\ 
   $Uniform_{[0.3,1]}$ &  
   0.66 & 0.09 & 11.06 \\ 
   $Mix_{Constant_{[0.75]},Uniform_{[0.3,1]}}$ & 
   0.70 & 0.05 & 21.18 \\ 
   $Mix_{Discrete_{[0:0.3,1:0.7]},Scores_{\{0.3,1\}}}$ & 
   0.68 & 0.15 & 6.64 \\ 
   $Random$ & 
   0.49 & 0.08 & 11.78 \\ 
   \hline
\end{tabular}
}
\caption{Synthetic models represented in Fig.~~\ref{fig:genDescription}.Each measure (the last three columns show performance, variance and regularity) is computed using 100 scores for each model (according to the pattern described in the first column). $Constant_x$: models with constant performance \inserted{$x$}; $Categorical_{[x:p_x,y:p_y]}$: models with a performance following a categorical distribution with values $x$ and $y$ and probabilities $p_x$ and $p_y$ respectively; $Uniform_{\{x,y\}}$: models with a uniform distribution of scores between $x$ and $y$; $Mix_{A,B}$: combination of models $A$ and $B$; $Random$: random scores between 0 and 1 (also using a uniform distribution).
} 

\label{Tab:models}

\end{table}

\begin{figure}[h]
	\centering
    \includegraphics[width=0.75\columnwidth]{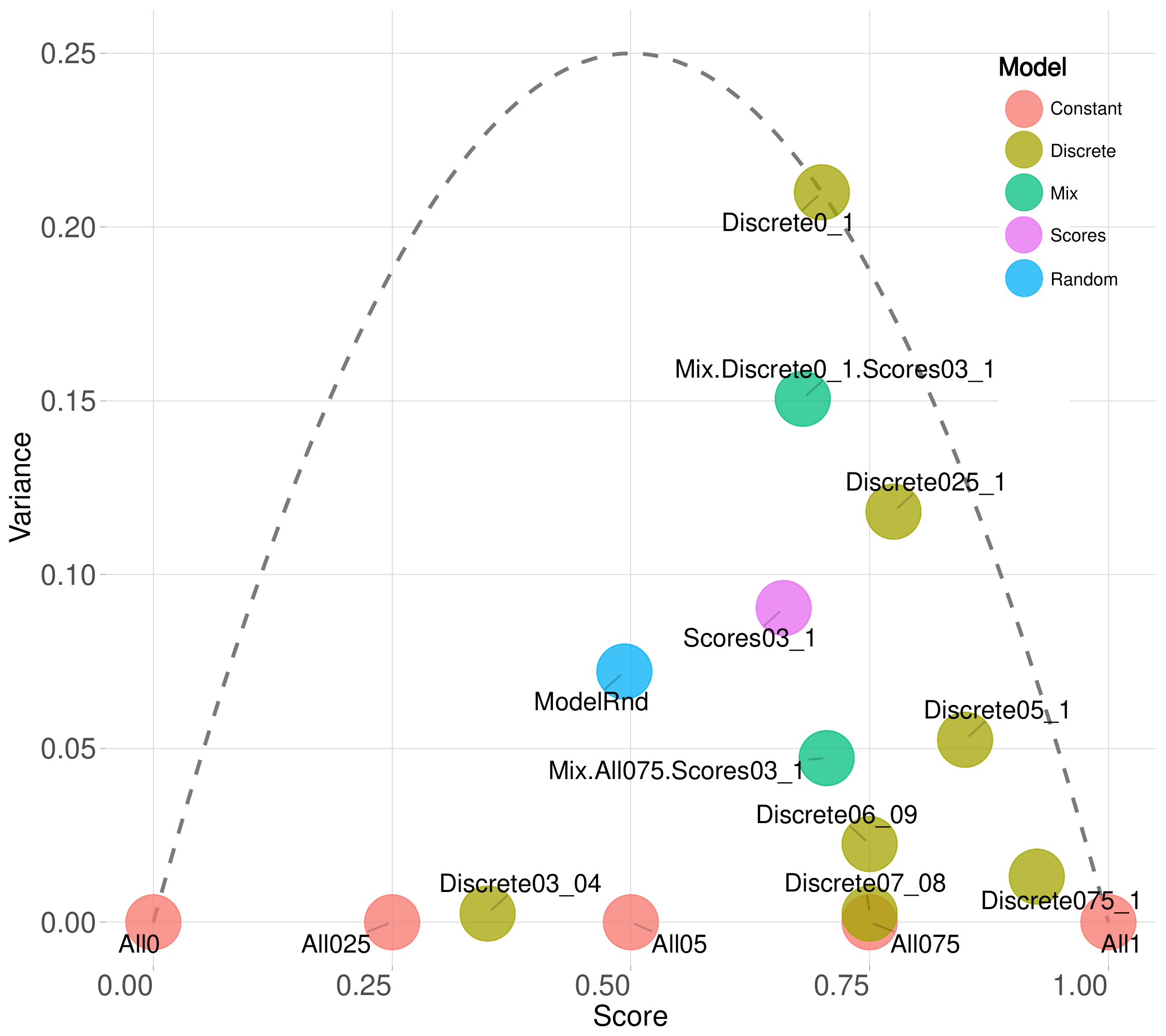}
    
    \caption{Variance of several synthetic models described in Table~\ref{Tab:models}.}
	\label{fig:genDescription}    
\end{figure}

Let us assume that performances are commensurate (so that we can average them) and that we can normalise performance between 0 and 1. Fig.~\ref{fig:genDescription} shows, on one hand, the maximum variance for a fixed average precision or score (normalised between 0 an 1) represented by a dashed grey curve (again the variance of the Bernoulli distribution). On the other hand, the average score (in the interval $[0,1]$) of several synthetic models (described in Table \ref{Tab:models}) are plotted against their variance for a set of 100 problems. The higher the models are, the less regular they are. We can see how those most regular methods (with lower variance in their results) are situated in the bottom part of the plot (closer to the \xaxis). As we see, the system that is always right (or the systems that is always wrong) has no variance, and perfect regularity.

But is this variance caused by failing at more difficult problems (as expected) or failing at some subfamilies (pockets) of problems? This is when difficulty comes in: any generality metric (and associated plot) has to be analysed in terms of difficulty. We now introduce a new metric of generality that does take difficulty into account:

\begin{equation}\label{eq:gen2}
\gendiff_j = \frac{1}{\sum_{h=1}^{h_{max}} {(\sigma^{[h]}_j)^2}}
\end{equation}

\noindent where $\sigma^{[h]}_j$ is the standard deviation of agent $j$ on all problems of difficulty $h$. 
Continuous difficulties could be handled by using an integral instead of a sum. However, as we will estimate generality from a sample, we assume difficulties are discrete (or discretised by bins). Note that for small samples the number of bins is important. For instance, if there is only one single bin for all items then generality becomes equal to regularity. The higher the number of bins the better, approaching a continuous notion of difficulty. However, in estimation, for a finite number of instances, binning has to be done with at least a minimum number of examples per bin, to avoid undefined or very unstable variances.

The estimation can be done up to a maximum difficulty $h_{max}$. Nevertheless, if we assume that systems have zero performance once a certain difficulty is reached, then there is no need to set a limit of difficulties on the sum. 

Given Equation~\ref{eq:gen2}, how can we get maximum generality? This is actually achieved when the slope of the agent characteristic curve is $-\infty$, i.e., the agent is perfect up to a given difficulty and hopeless from that moment on. In this case all variances are 0 and generality is infinity. 

This relation of generality to the slope of the agent characteristic curve finally completes the circle and the duality between agents and problems, since generality can be seen as dual to discrimination (the slope of the item characteristic curve). For binary answers, there is no extra degree of freedom, and this extra parameter is confounded by all the others. But for continuous values (scores or probabilities), the slope has this extra degree of freedom. 

In the same way that two agents with the same average performance (or ability parameter) cannot be distinguished as more or less general, we can have two systems with the same generality value with very different behaviour. For instance, we can have an agent A that is perfect up to difficulty $h=5$ and an agent B that is perfect up to a difficulty $h=2$. Looking at generality, both would have $\gamma=\infty$ and would then be indistinguishable with generality. But, clearly, agent A is more capable than agent B. Actually, in this case, we have a very interesting way of looking at their relation. If two agents are perfectly general and one is more capable than the other, we have a dominance relation. In our example, A dominates B. We can assert that whatever is solved by B is solved by A. This is closely related to the intuition of the transitivity of performance, already explored in previous papers of the GVGAI competition, which has raised doubts about the generality of the participants \cite{nielsen2015towards, bontrager2016matching}. 

In the following sections, we will better analyse the behaviour and interpretability of this new notion of generality. But let us first introduce the two benchmarks we will work with.

\section{Benchmarks: ALE and GVGAI}\label{Data}
In this section we will describe the benchmarks (ALE and GVGAI) we will use for the experiments in the next section\footnote{For the sake of reproducibility and transparency, all the code and data is on Github (\url{https://github.com/nandomp/AI_benchmark_analysis}).}.

\subsection{The Arcade Learning Environment}

The Arcade Learning Environment (ALE) was introduced by \cite{bellemare2015arcade}, after compiling a good number of games \modified{for Atari} 2600, a popular console of the late 1970s and most of the 1980s. The simplicity of the games from today's perspective and the use of a visual input of 210 $\times$ 160 RGB pixels at 60Hz makes the benchmark sufficiently rich (but still simple) for the AI algorithms of today. After  \cite{mnih2015human} achieved superhuman performance for many of the ALE games, the benchmark became very popular in AI. There are so many platforms, techniques and papers using ALE today that the results on this benchmark are usually analysed when talking about breakthroughs\footnote{\url{http://cdn.aiindex.org/2017-report.pdf} and \url{https://www.eff.org/ai/metrics}} and progress\footnote{\url{http://www.milesbrundage.com/blog-posts/my-ai-forecasts-past-present-and-future-main-post}} in AI.

We have performed a bibliographical search to find all the papers that include experiments with a wide range of ALE games. We first discarded those techniques that use look-ahead access to a simulator (this is common in search-based approaches \cite{naddaf2010game,lipovetzky2015classical,shleyfman2016blind}, but not comparable to humans). This is due to the real-world situation of human players that have to perform with no access to the game other than the screen, and this is the standard for comparison. Look-ahead techniques could have been studied separately (the ones we use in the next section with GVGAI), but what should not be done is to combine results with different rules.

Hence, we will use the results obtained with truly learning approaches (most, but not necessarily all, using reinforcement techniques, usually in conjunction with deep learning). In this category, we are flexible about whether the results include human demonstrations or not (``noop'' and ``humanstarts'' settings). Overall, we integrated about 40 techniques from about a dozen papers covering classical deep reinforcement learning techniques (DQN) \cite{mnih2013playing,mnih2015human,furelos2015learning}, as well as specific adaptations to the DQN such as those using {\em duelling} architectures \cite{wang2015dueling}, those with prioritised experience replay \cite{schaul2015prioritized}, or those reducing inherent estimation errors of learning \cite{van2016deep}. We also analysed more recent approaches which improve the stability, convergence and runtime of DQN \cite{he2016learning,o2017combining,pritzel2017neural,talvitie2015pairwise}, as well as some distributed/parallel versions \cite{gruslys2017reactor,nair2015massively}. Evolution strategies \cite{salimans2017evolution}, such as a scalable alternative to DQN, were also included. We discarded some papers because they did not include the results for all the 49 games that are most common in many papers. As some results (especially DQN) are reported repeatedly for some papers, we removed all results with a correlation higher than 0.99. In other cases, the results for the same technique with different parameters were kept. We also removed repeated results. Note that some techniques, such as DQN, are used repeatedly, but with different conditions (parameters). Only exact equal results were removed.

\subsection{The General Video Game Playing Competition}

The General Video Game AI (GVGAI) competition \cite{perez20162014} was one of the first AI competitions featuring a significant number of unseen games within a relatively large problem space  (after \cite{genesereth2005general}). As in ALE, this competition focuses on video games, in particular two-dimensional games including classic arcade, puzzles, shooters and many more. The games can also differ in the way players are able to interact with the environment (actions), the scoring systems, the objects that are part of a game or the conditions to end the game.  Unlike ALE, GVGAI was created to avoid participants tailoring their submissions to a few well-known games. Instead, participants are pitted against a number of unseen games. Another difference is that controllers are able to access an abstract  representation of the game state (so complex perception is not needed) as well as a simulator so that (look-ahead) tree search algorithms and other planning approaches can be used.

Because of this access to the simulator, those controllers based on Monte Carlo Tree Search (MCTS) \cite{browne2012survey}, Rolling Horizon Evolutionary Algorithms (RHEA) \cite{perez2013rolling} as well as hybrids with popular tree search methods  have been successful on this benchmark. However, as shown in \cite{nielsen2015towards}, performance is non-transitive since different controllers play different games best and, thus, no algorithm dominates all the others.

Regarding the data, we will work with the scores of 49 games and the 23 controllers (agents) that were submitted to the 2015 GVGAI competition\footnote{Results courtesy of Julian Togelius.} \cite{bontrager2016matching}. Each game has 5 levels, and each level was attempted 5 times. This makes a total of $23 \times 49 \times 5 \times 5 = 28175$ trials. For each trial the data includes the win/loss achieved by the controller.

\subsection{Normalisation}

In the case of ALE, we have point scores (usual in videogames), which are clearly not commensurate (10,000 points in a game may be low while 50 in another may be high). It is then common to normalise them by human scores (where 0 equals random, and 100 equals human level), usually putting human level as a target for a ``successful'' or ``acceptable" result. On the other hand, for GVGAI 
we do not have human results as a reference, but a notion of success is given by the ``win/loss'' values ($1$:win, $0$:loss), which indicate whether the agent `beat' the game. Each game was attempted 5 times, so win/loss values can be averaged by the number of trials to obtain scores. When analysing both benchmarks, we have linearly scaled their results to $z$-scores (or standard scores) for both benchmarks so that we can compare the results in a more meaningful way. Then we apply the error function, so $\Response_{j,i}$ always ranges from 0 to 1.

\section{Task analysis: difficulty and discrimination} \label{TaskAnalysis}

\begin{figure*}[h]%
    \centering
    \includegraphics[width=0.49\columnwidth]{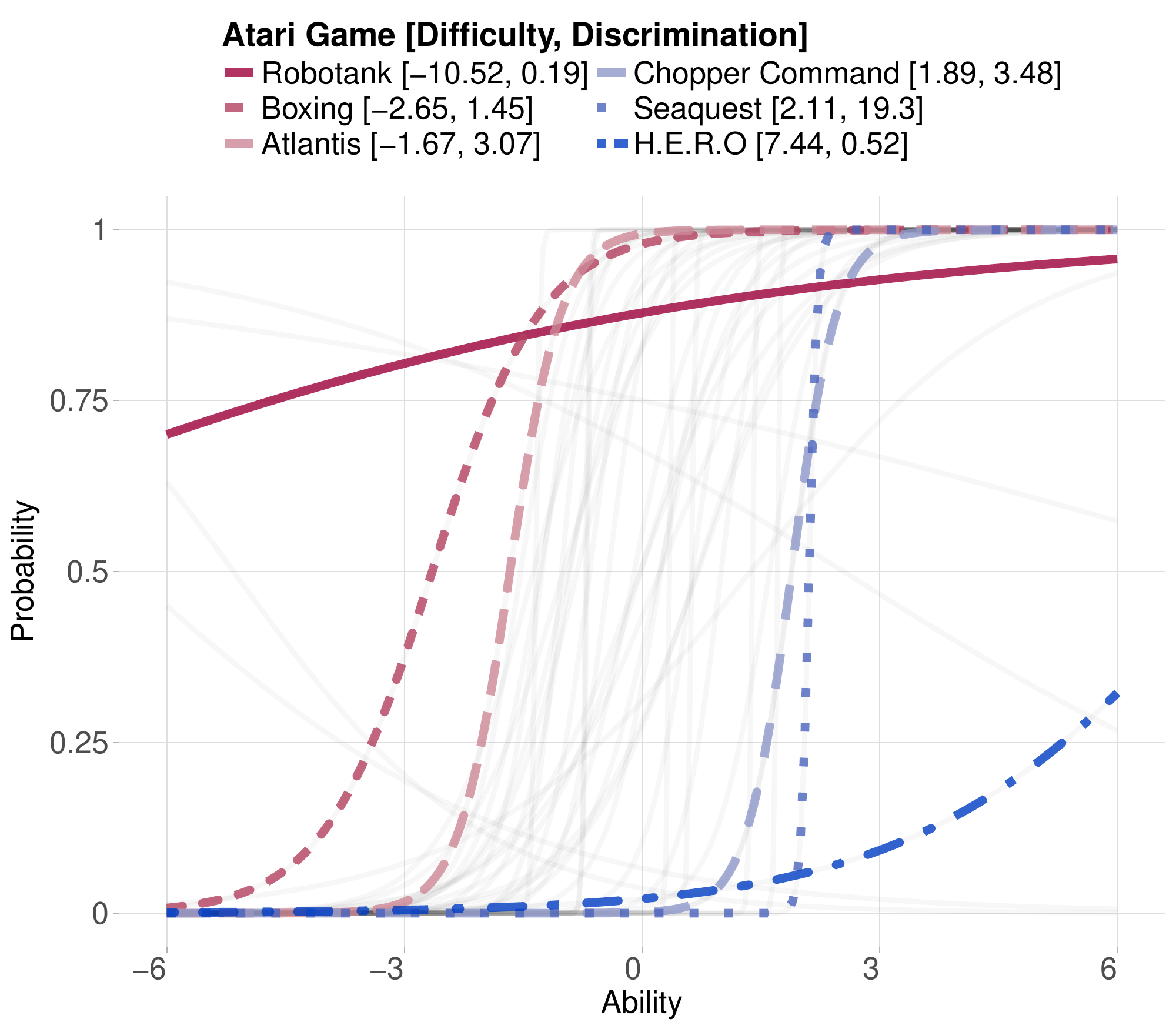}
    \includegraphics[width=0.49\columnwidth]{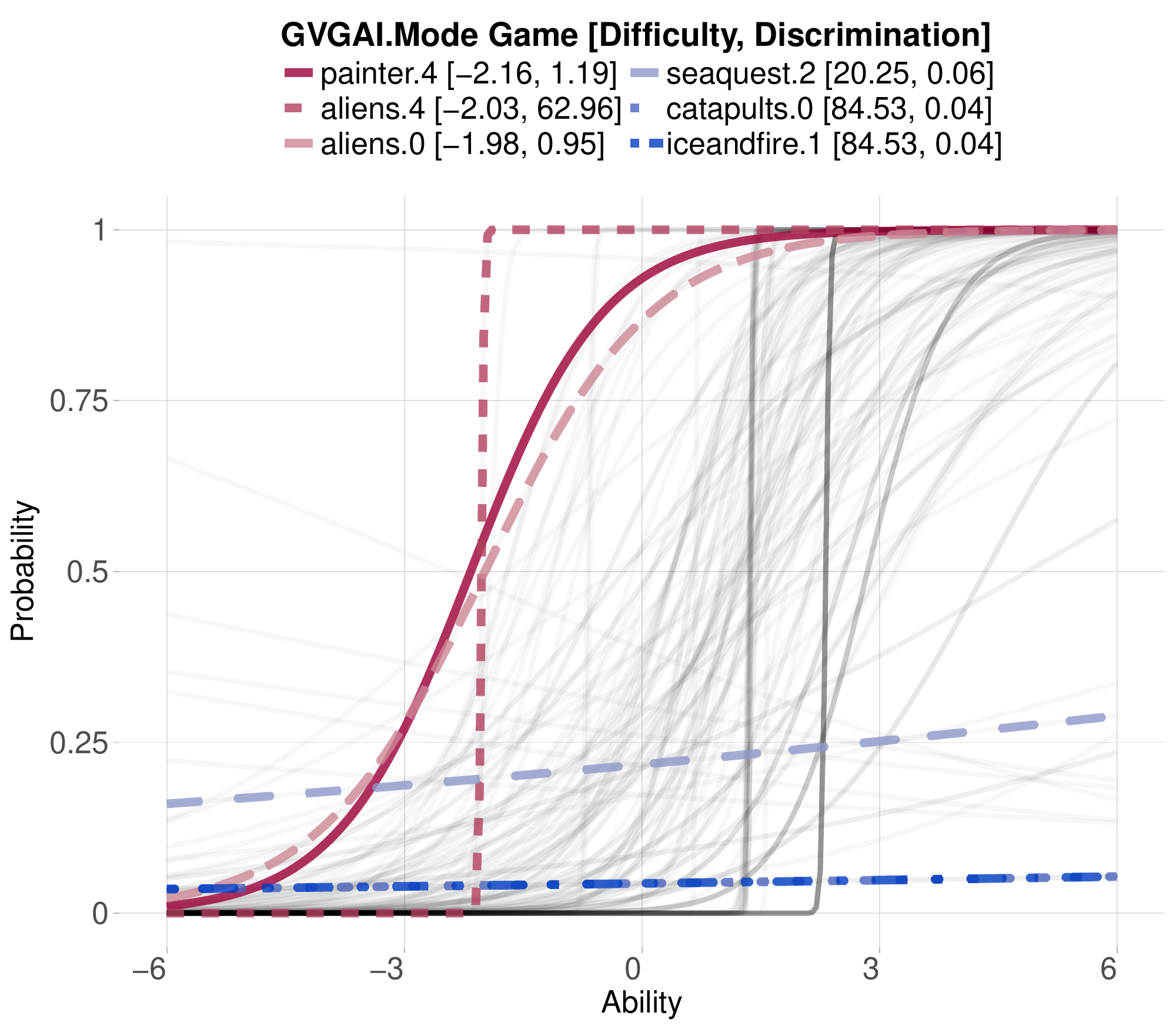}
    \caption{ICCs of the 3 most (bluish colours) and least (reddish colours) difficult ALE games (left) and GVGAI games (right). Negative discrimination instances filtered out. 
    All ICC plots from both benchmarks are shown in grey. 
    }%
    \label{fig:mostDiff}%
  
\end{figure*}

\begin{figure*}[!h]%

    \centering
    \includegraphics[width=0.49\columnwidth]{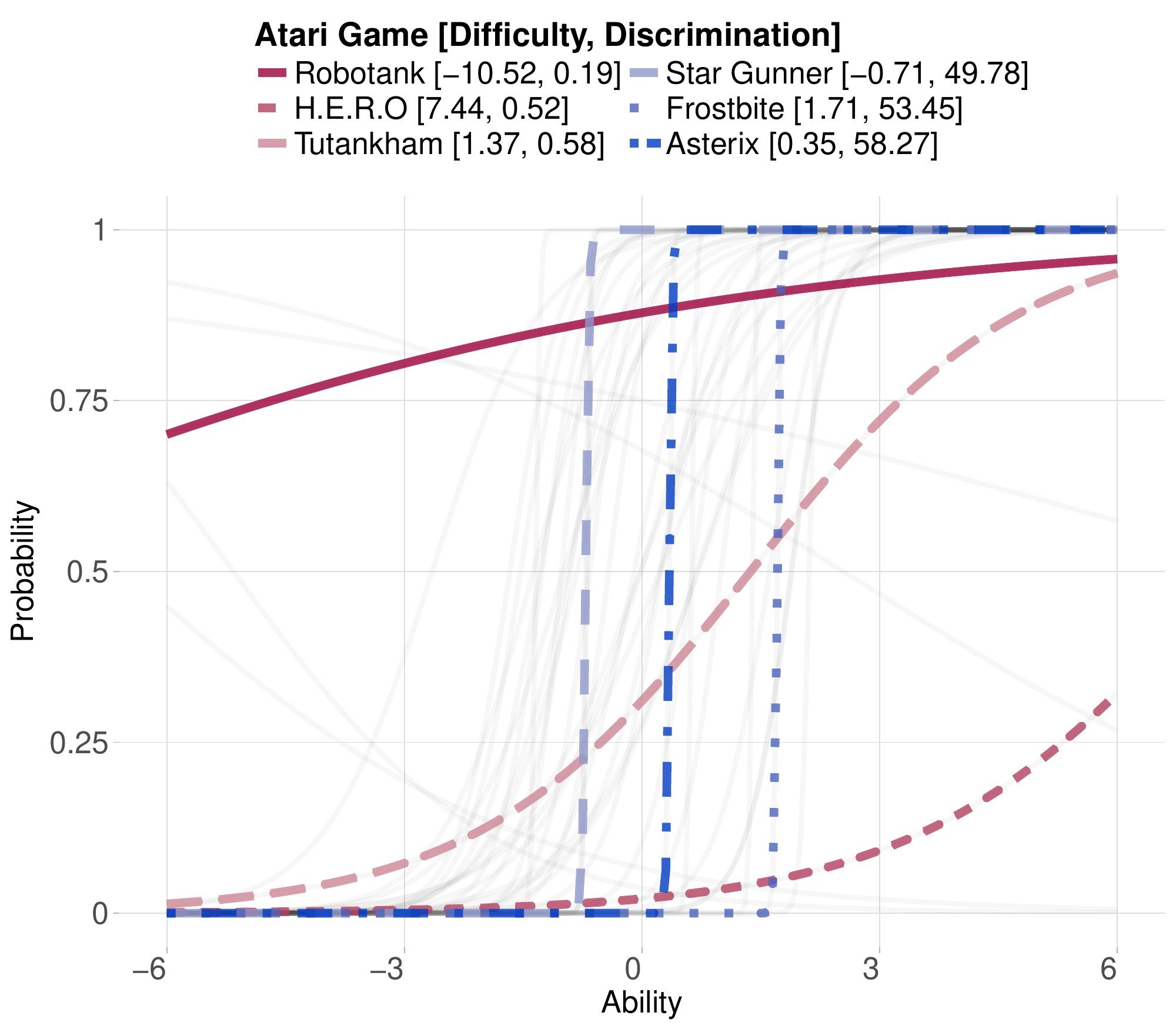} 
    \includegraphics[width=0.49\columnwidth]{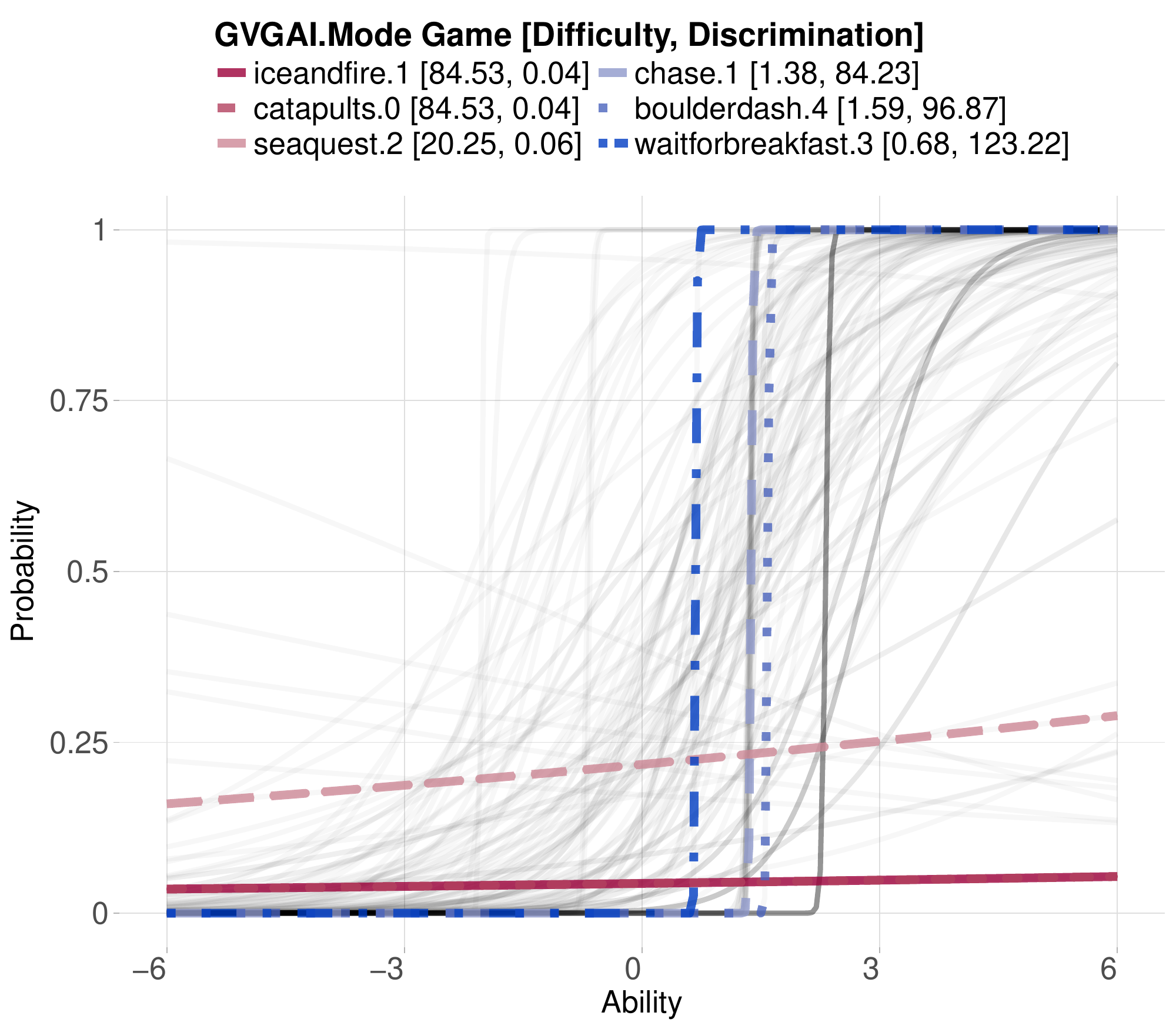}%
    \caption{ICCs of the 5 most (bluish colours) and least (reddish colours) discriminating ALE games (left) and GVGAI games (right). Negative discrimination instances filtered out. All ICC plots from both benchmarks are shown in grey.}%
    \label{fig:mostDisc}%
\end{figure*}

\begin{figure*}[!h]%
    \centering
 	\includegraphics[width=0.49\columnwidth]{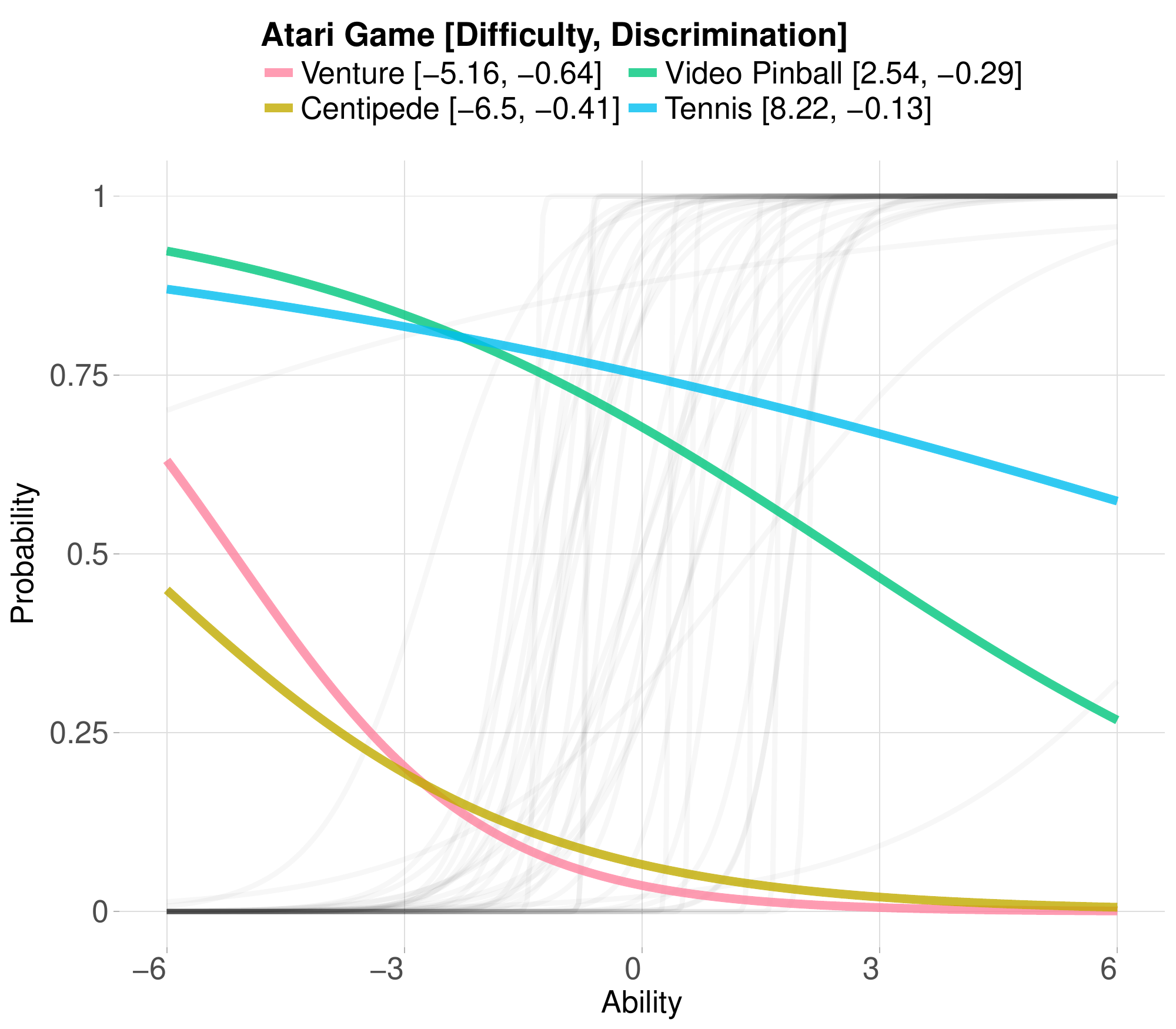} 
    \includegraphics[width=0.49\columnwidth]{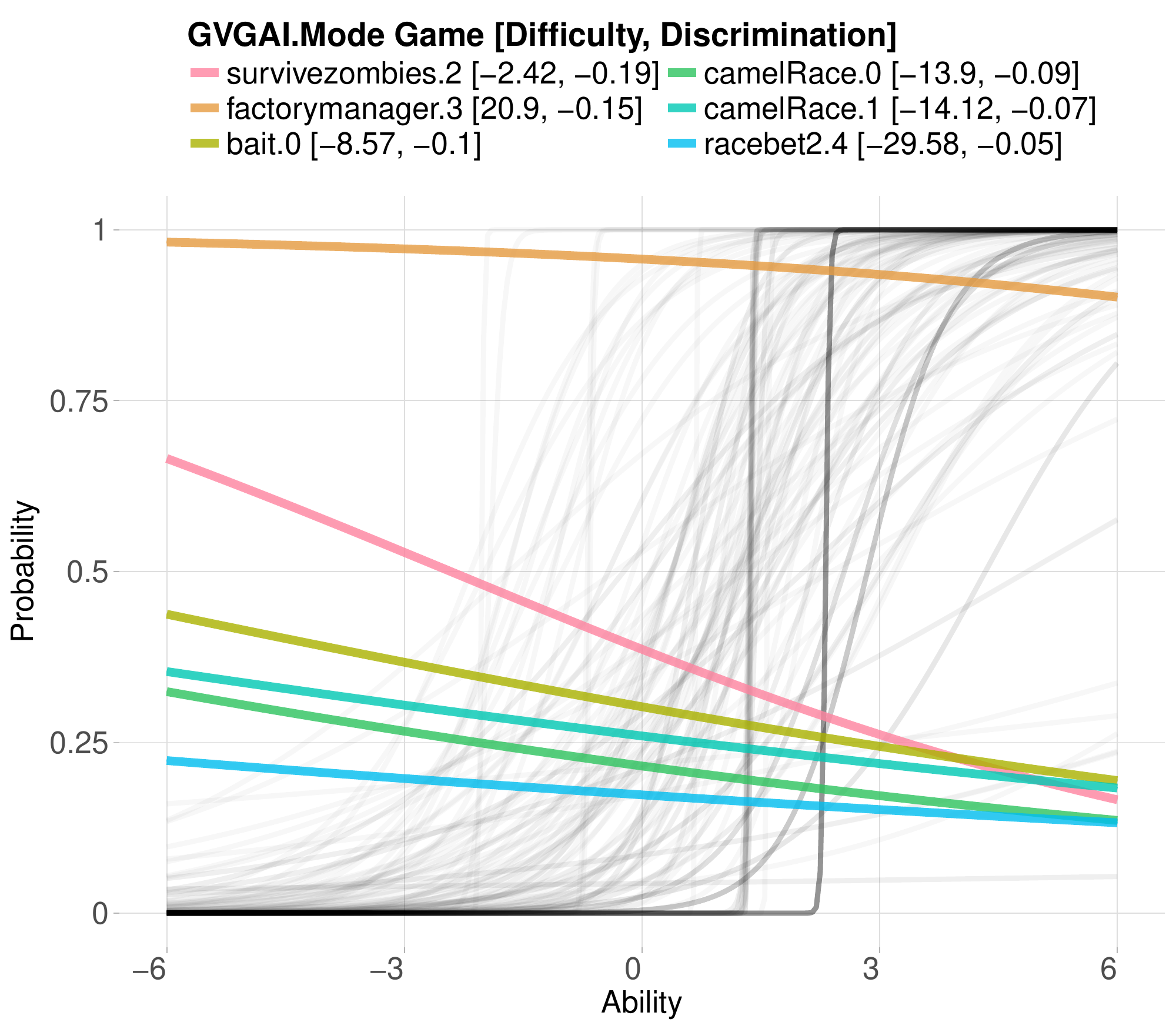}%
    \caption{Examples of ICCs of those ALE (left) and GVGAI (right) games with negative discrimination. Technique abilities are also included in the ICCs, plotted at $y = 1$ if their score is above 100, and at $y = 0$ otherwise.}%
    \label{fig:negativeICC}%
\end{figure*}

Even if we normalise or scale the scores, this does not give us any idea of the difficulty of the tasks or their discriminating power. In these games it is hard to derive a specific notion of difficulty in terms of the features of the games. The number of NPCs, the size of the game or other metrics would not help much to a notion of difficulty\footnote{The ALE games include the notion of \textit{mode}, which alters the games (e.g, changing  the  game  
dynamics,  actions, etc.), but it is not linked to difficulty. ALE supports non-commensurate difficulties since its version 0.6 (Sept 2017), but no new systematic results 
are available with these.}. Other general definitions of theoretical difficulty and discrimination \cite{orallo2017} would be computationally very expensive. As a result, a populational approach, as represented by IRT, seems the most straightforward approach to derive the difficulty and the discrimination parameters. In order to apply binary IRT, we consider a `success' as explained in the previous section (above or equal human performance for ALE, and  equal or more wins than loses for GVGAI games).

Some tasks for ALE were always below (\textit{Alien}, \textit{Asteroids}, \textit{Bowling}, \textit{Gravitar}, \textit{Montezuma}, \textit{Ms. Pacman} and \textit{Private Eye}) or always above (\textit{Krull}) human performance  for all techniques (i.e, constant results after normalisation). 
This implies that the IRT models cannot be fitted for these tasks, so these  were excluded for the rest of the analysis.  Similarly, some tasks for GVGAI (\textit{Bait}, \textit{Bolo adventures}, \textit{Camel Race}, \textit{Factory manager}, \textit{Firestorms}, \textit{Modality}, \textit{Portal}, \textit{Real Portal}, \textit{Realsokoban}, \textit{The citadel} and \textit{Wait for breakfast} 
for specific $modes$) were removed as they have the same result (0 or 1) for all agents.

Once the data is ready, a 2-parameter IRT logistic model (2PL) is learned for each ALE and GVGAI game. We adopt MLE to estimate all the model parameters for all instances and the classifier abilities simultaneously, as usual in IRT. In particular, for generating the IRT models, we used the {\ttfamily{ltm}} R package\footref{ltm}, using Birnbaum's method, as explained in section~\ref{IRT}. {The package \ttfamily{ltm}} (as many other IRT libraries) outputs indicators about the goodness of fit, which can be used to quantify the discrepancy between the values observed in the data (items) and the values expected under the statistical IRT model. Item-fit statistics may be used to test the hypothesis of whether the fitted model could truly be the data-generating model or, conversely, we expect the item parameter estimates to be biased. In practice, an IRT model may be rejected on the basis of bad item-fit statistics, as we would not be reasonably confident about the validity of the inferences drawn from it \cite{maydeu2013goodness}. Apart from the goodness of fit, in order to double-check the results, we recommend re-estimating the parameters with different initial values (seeds) for every model you fit in order to check whether the estimates are consistent. In the present case, none of the estimated models were discarded because of bad item-fit statistics or inconsistency in their results.

Regarding the results, for the ALE games, difficulties range from $-10.51$ to $8.22$, while discriminations range from $-0.64$ to $58.27$. For the GVGAI games, difficulties range from $-29.58$ to $84.53$, while discriminations range from $-0.19$ to $123.22$.

The item parameter that is easiest to understand is difficulty. Because of the MLE estimation method, the value is not equal but well correlated with the percentage of AI techniques that are successful for the game. Intuitively, easy games are solved by almost all techniques, and difficult games are those that are only solved by very able techniques.  Fig.~\ref{fig:mostDiff} shows the ICCs of those three most (and least) difficult ALE (left) and GVGAI (right) games with positive discrimination. From those games, the most difficult ALE game seems to be \textit{H.E.R.O}, and \textit{iceandfire.1} for GVGAI. However, we  see cases such as \textit{Tennis} (ALE), which has the highest difficulty ($8.22$) but negative discrimination ($-0.13$, Fig.~\ref{fig:negativeICC} left). According to \cite{bellemare2015arcade}, it is a challenging game that requires fairly elaborate behaviour before observing any positive reward, but simple behaviour can avoid high negative rewards by not ever serving, which possibly explains the negative discrimination. Something similar happens with the GVGAI games, where \textit{factorymanager.3} is the third  most difficult one ($20.9$), but its discrimination is negative ($-0.15$, Fig.~\ref{fig:negativeICC} right). 

The discrimination parameter (slope) measures the capability of a game to differentiate between techniques. Therefore, when applying IRT to evaluate techniques, the slope of an instance can 
distinguish between strong or weak techniques. Fig.~\ref{fig:mostDisc} shows the ICCs of the most discriminating ALE (left) and GVGAI (right) games. From the 41 ALE games analysed, 37 had positive discrimination. Regarding the 154 GVGAI games of different modes analysed, 148 had positive discrimination. For all these the probability of correct responses is positively related to the estimated ability of the techniques. However, negative discriminations were observed for 4 ALE games (Figure \ref{fig:negativeICC} left)  and 6 GVGAI games (Figure \ref{fig:negativeICC} right).

These ``abstruse" cases (most frequently solved  by the weakest techniques) are anomalous in IRT,  
and should be considered with extreme care for the analysis of new AI algorithms. Are these games particularly difficult or are they just useless for evaluation since most able techniques do worse than those less able ones? Should we restrict our benchmarks to those items to positive (preferably high) discrimination? That depends on the purpose and resources of the evaluation, but what is clear is that, in order to determine which games are most informative for the analysis of new AI algorithms, difficulty alone is insufficient: we also need to look at discrimination.

\begin{figure}[ht]%

    \centering
    \includegraphics[width=0.49\columnwidth]{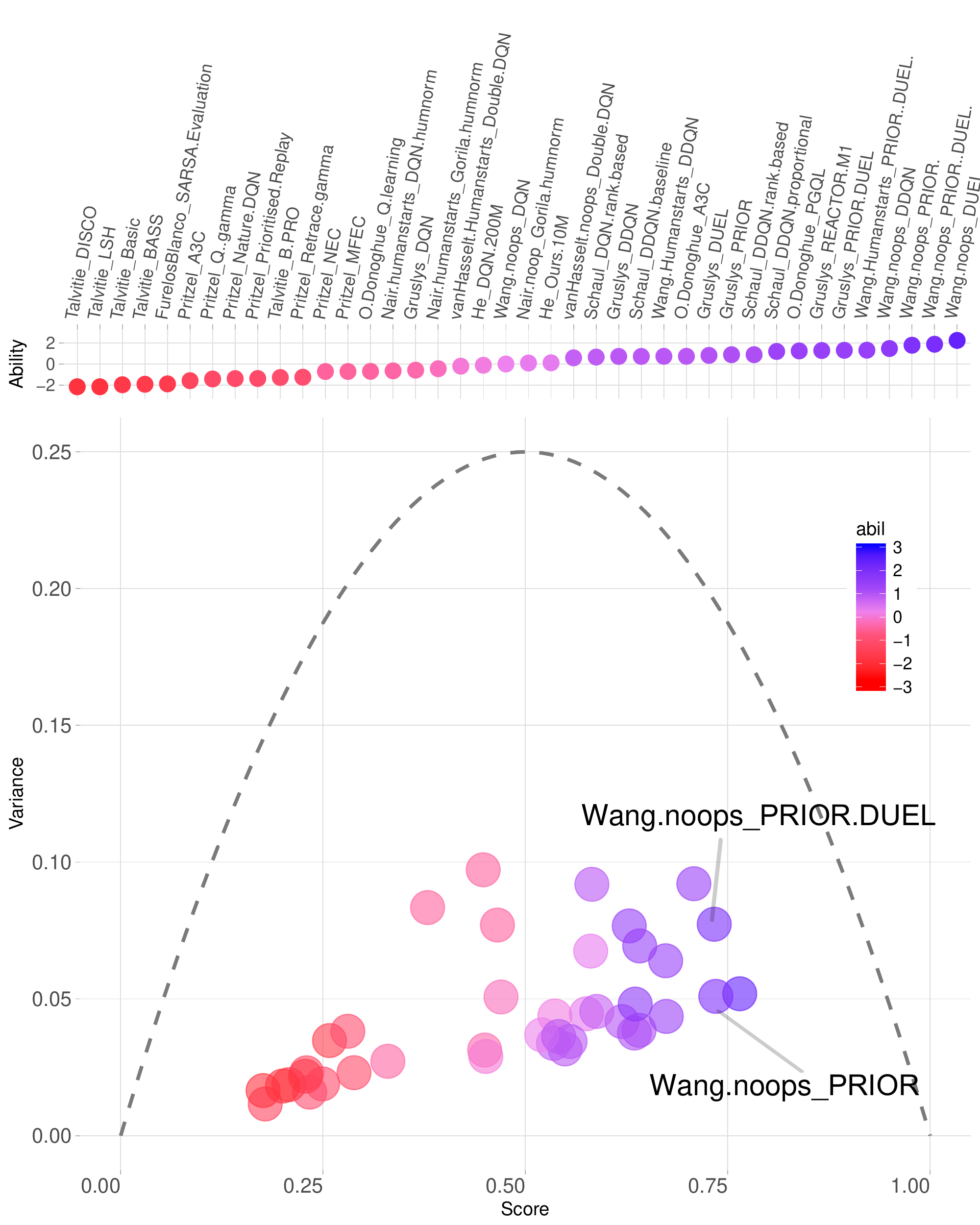}  
      \includegraphics[width=0.49\columnwidth]{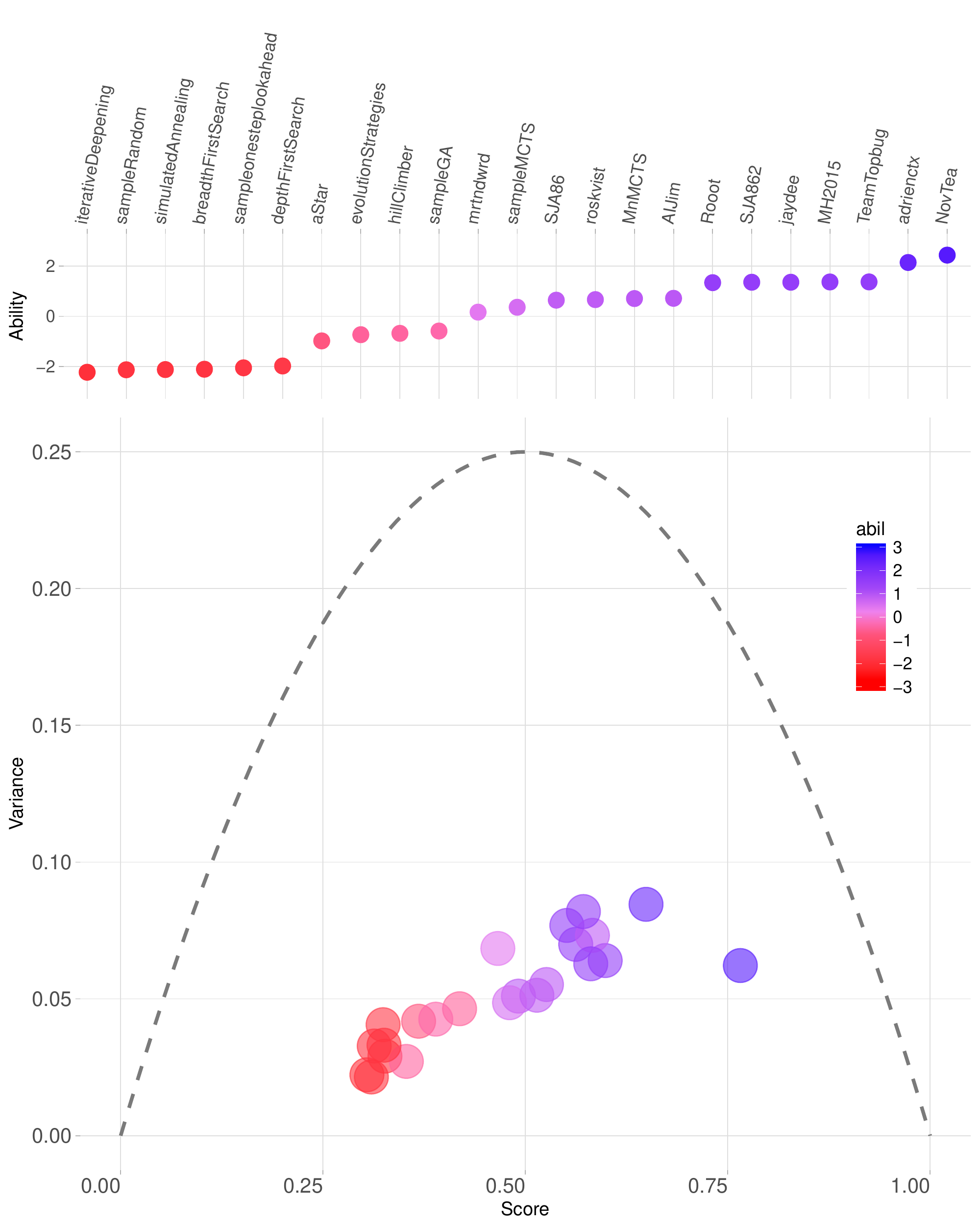}  %
    \caption{Variance vs. average normalised score for the AI techniques included in this study (left: ALE, right: GVGAI). The IRT abilities are shown with graded colours from red to blue. The dashed grey curve is the variance of a Bernoulli distribution (the worst case).}%
    \label{fig:variance}%

\end{figure}

In a nutshell, the discrimination parameter provides us an extra dimension to characterise a game. If a game has positive discrimination, it is actually well aligned with ability, and only the good AI techniques obtain good scores. In case a game has a discrimination close to 0, it has a high failure ratio (but happens with good and bad classifiers equally). Finally, if a game has negative discrimination, it is not aligned with ability (with more good AI techniques failing to obtain good scores than bad classifiers). These three cases explain the role of the discrimination parameter when evaluating different AI techniques.

\begin{figure*}[!ht]
  {\includegraphics[width=0.49\columnwidth]{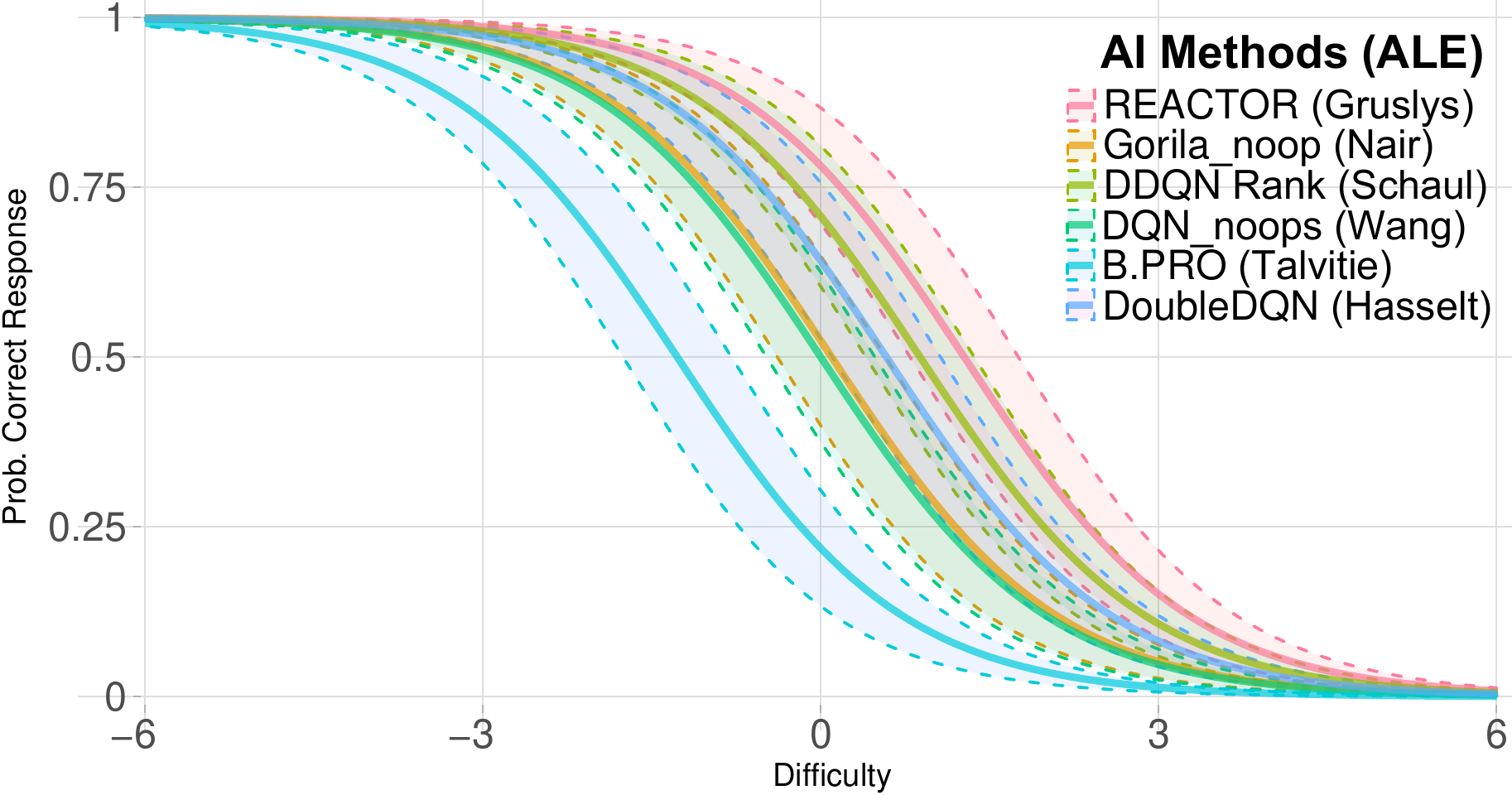}
  \includegraphics[width=0.49\columnwidth]{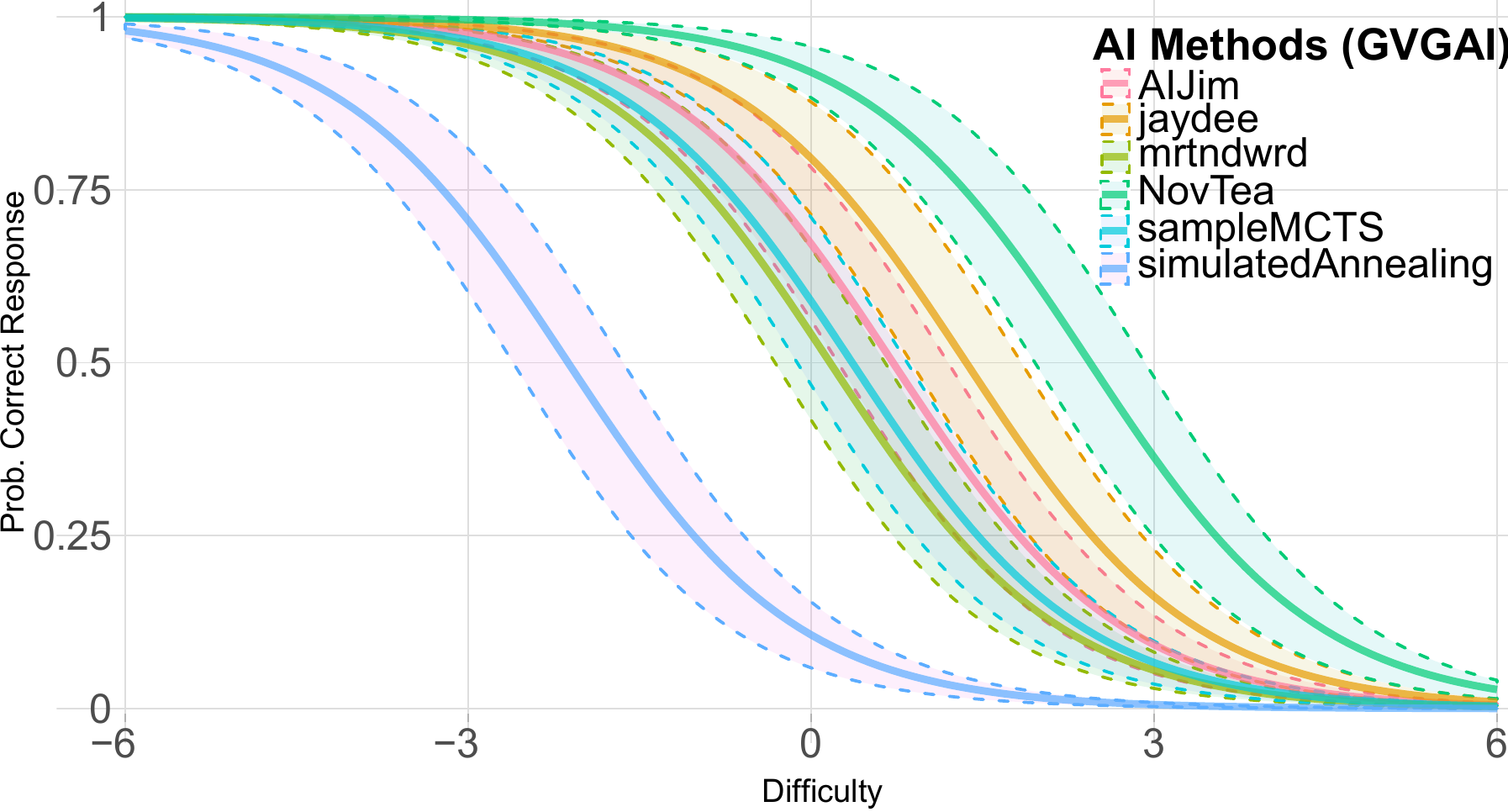}\label{fig:sub2}} \\[\baselineskip]%

{\includegraphics[width=0.49\columnwidth]{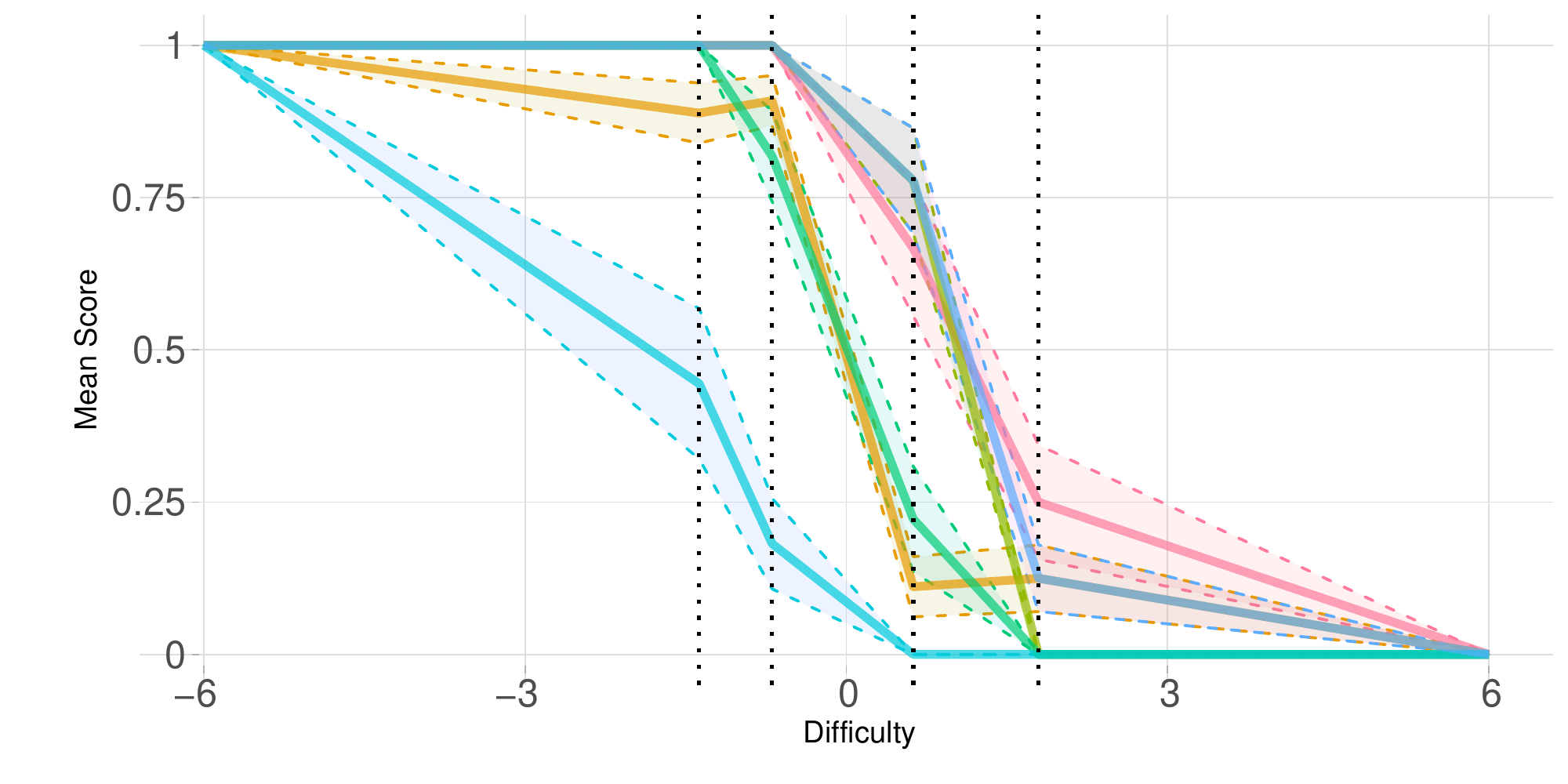}
  \includegraphics[width=0.49\columnwidth]{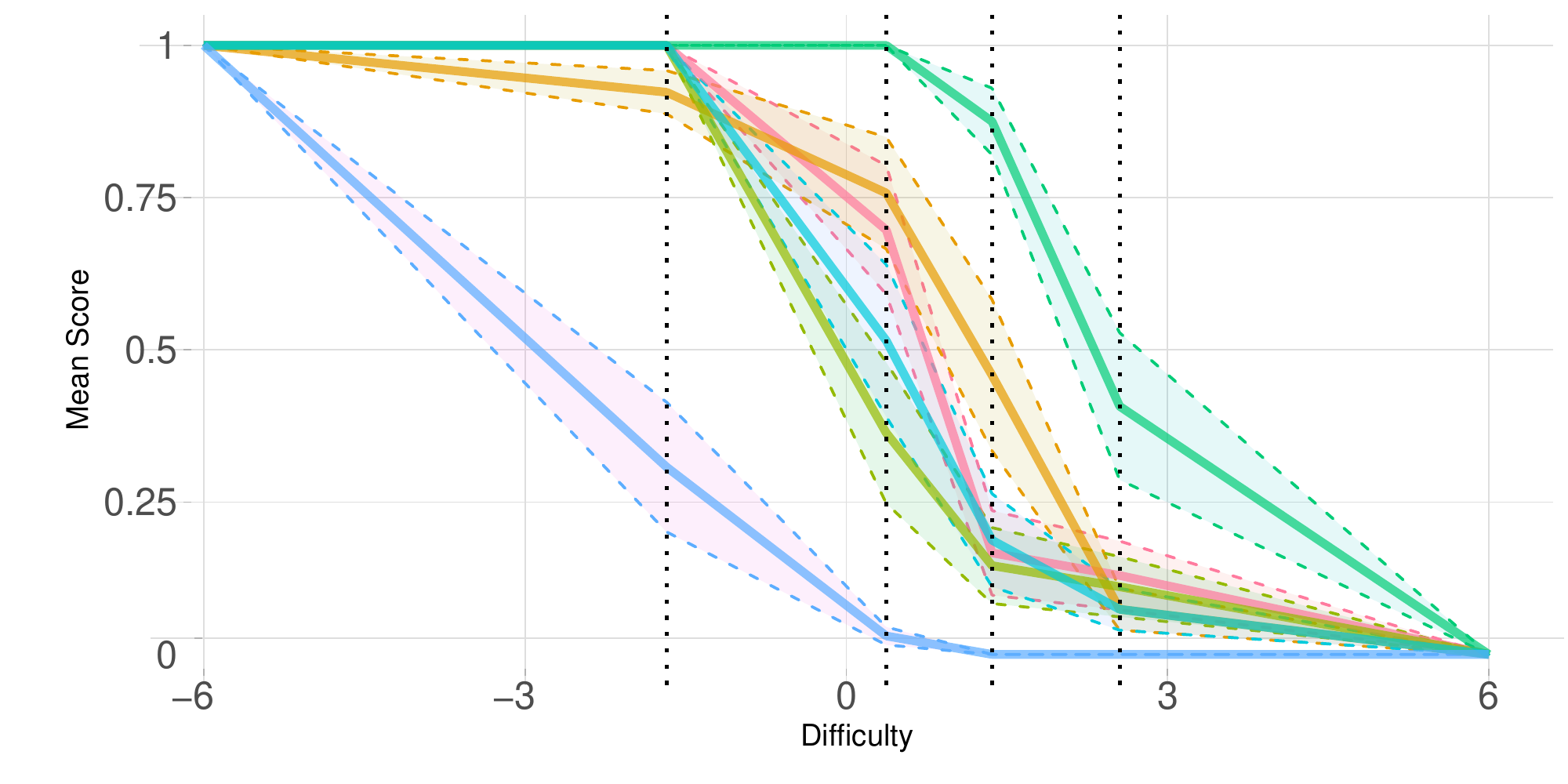}\label{fig:sub1b}}%
  \caption{$(top)$ Theoretical ACCs, i.e., probability of a correct response as a function of the difficulty parameter for an illustrative subset of techniques for ALE (left) and GVGAI (right). 
  $(bottom)$ Empirical ACCs (across bins on the difficulty parameter) for the same subset of techniques. In order to see some progression in the curves (sufficient detail) but still some robustness without spurious peaks, the bins had to contain a minimum number of instances in each interval. Consequently, we set a minimum number of 4 bins with at least 10 examples per bin. Dashed black vertical lines represent the average difficulty values for the instances in each bin. Variance is also represented for each technique (semitransparent ribbon in $\pm(\nicefrac{1}{2})\sigma^2$). 
  Negative discrimination instances filtered out.}\label{fig:test}
  
\end{figure*}

\section{Technique analysis: ability and generality} \label{TechAnalysis}

As we mentioned in Section \ref{IRT}, IRT has a dual character: we get information about the items (games) but also about the respondents (AI techniques). IRT estimates a value of ability $\theta$ for each AI technique. 
Unlike average scores, ability takes difficulty into account and is normalised.  For instance, if an AI system scores well for difficult games but fails for some easy ones, IRT can give it more value than the opposite situation, depending on their item parameters. Also, IRT penalises those AI techniques that score well in games with negative discriminations.

If there are not many items with negative discrimination, as in our case, ability will be similar to an aggregation of results.  Fig.~\ref{fig:variance} shows scores on the \xaxis and ability as graded colour from red to blue, with almost perfectly aligned rankings.

Ability assumes that agents are better at easy instances than they are at difficult instances, but this implies an {\em uneven} treatment on subpopulations of problems. As discussed in previous sections, this is reasonable. The question is what pattern the ``unevenness" (or dispersion) has. If it is actually unrelated to difficulty, we may have pockets of good performance (and pockets of poor performance) with different problem patterns, and the technique would not be very general.

Let us start with the global variance as an indication of regularity, not taking difficulty into account. In Fig.~\ref{fig:variance}, we see that for the same score and ability value, regularity may vary significantly. For instance, for ALE, {\ttfamily{PRIOR DQN}} and {\ttfamily{PRIOR.DUEL}} (DQN-based methods from \cite{wang2015dueling}, labelled in Fig.~\ref{fig:variance}, top) have similar score and ability, but the former seems more regular (with a variance of only 0.05).

But is this actually a measure of generality? Is it so different from the definition of generality in equation \ref{eq:gen2} that does take difficulty into account? Let us find out. Fig.~\ref{fig:test} (top) shows the theoretical agent characteristic curves (ACCs). The IRT models are logistic, with just one parameter varying for the agents (the position, i.e., the ability). 
If we fix the difficulty, and assume the discriminations $a_h$ are similar for all difficulties $h$, we have that all slopes are the same. That means that for each technique $j$, using equations \ref{eq:3pl} and \ref{eq:gen2}, and applying the variance of a Bernoulli distribution, we have:

\begin{eqnarray*}\label{eq:gen3}
\gendiff_j 
& = &  \frac{1}{\sum_{h=1}^{h_{max}} {\frac{1}{1+e^{-a_h(\theta_j-h)}}(1-\frac{1}{1+e^{-a_h(\theta_j-h)}})}} \\
& = &  \frac{1}{\sum_{h=1}^{h_{max}} {\frac{e^{-a_h(\theta_j-h)}}{(1+e^{-a_h(\theta_j-h)})^2}}} 
\end{eqnarray*}

\noindent We would have that if $\theta_j$ is sufficiently large (so that the variance approaches zero for $h=1$) and not sufficiently close to $h_{max}$ (so that the variance approaches zero for $h=h_{\max}$), then we would have the same $\gamma_j$ for each $j$, as the only thing that changes is location. This is what we see in Fig.~\ref{fig:test} (top).

\begin{figure}[h]%

    \centering
    \includegraphics[width=0.4\columnwidth]{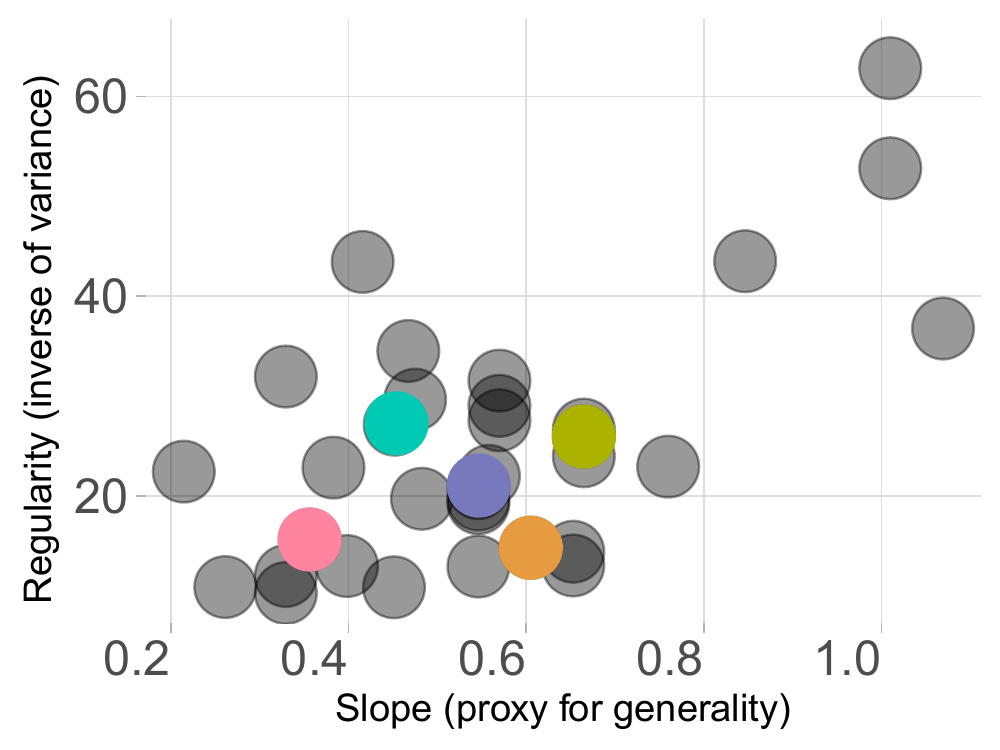} 
    \includegraphics[width=0.4\columnwidth]{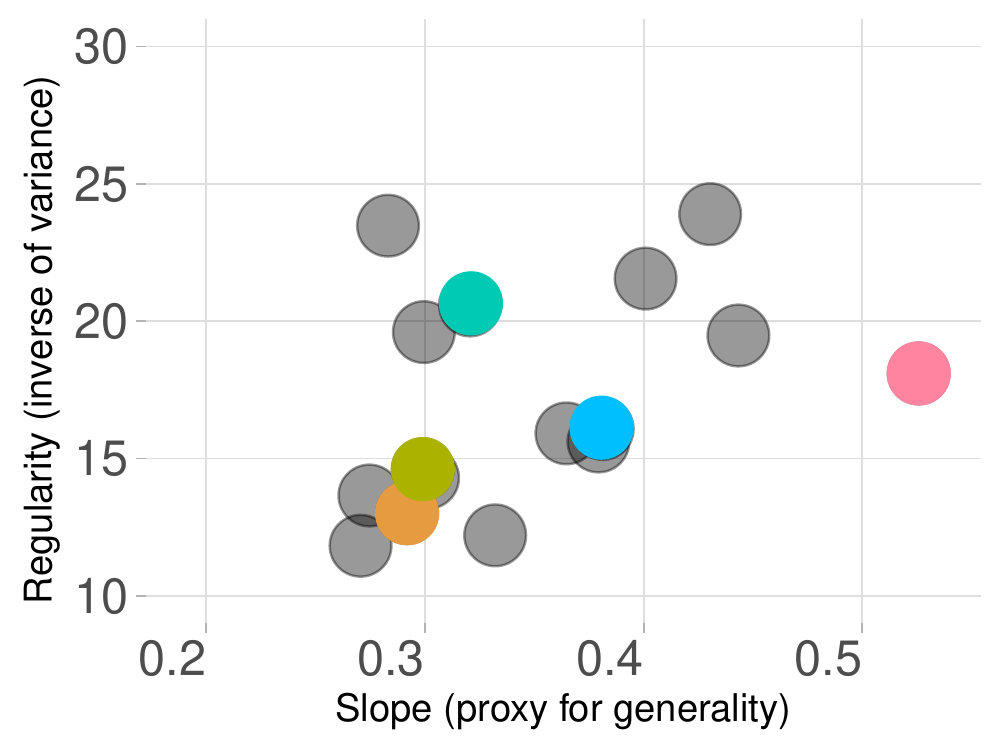} 
    \caption{Slope (proxy for generality) against regularity (inverse of the variance) for all the analysed techniques. Left: ALE, right: GVGAI.  Coloured points represent those systems in Fig.~\ref{fig:test}.}
    \label{fig:vargen}%

\end{figure}

However, the empirical curves, as shown in Fig.~\ref{fig:test} (bottom) give us a different view. Some techniques have different slopes, which, together with the discontinuities and monotonicities, give us different generalities, as shown with the areas of the semitransparent ribbons (the smaller the area inside the ribbon at the bin points the higher the generality). See, for instance, the methods {\ttfamily {DDQN Rank}} (ALE, in olive green colour) or {\ttfamily {AIJim}} (GVGAI, in pink colour) in Fig.~\ref {fig:test} (bottom). Both have high values of generality but low regularity. The reverse is also the case. {\ttfamily {NovTea}} (GVGAI, in green marine colour) has high regularity but medium-low generality value. All this can be seen more clearly in Fig.~\ref{fig:vargen}, where we use the slope of the empirical curves at mean score 0.5 as a proxy for generality (the higher the slope the higher the generality).

These observations are somehow confirmed by the intrinsic nature of the methods used. If we focus on the GVGAI methods because of its wider variety of algorithms, we see that {\ttfamily{{AIJim}}}, a variant of  \emph{MCTS} that performs well in several related domains (see \cite{browne2012survey}), has high generality. As for the low generality of {\ttfamily{NovTea}}, it is an \emph{Iterated Width}-based approach \cite{lipovetzky2012width}, originally a planning technique, which tries to outperform MCTS in GVGAI with {\em specific} tuning (pruning using novelty test) \cite{bontrager2016matching}.

Fig.~\ref{fig:vargen} shows there is a correlation between regularity and generality ($0.62$ and $0.4$ for ALE and GVGAI, respectively), but they are different concepts. To see this more clearly, we find a negative correlation between ability and regularity ($-0.68$ and $-0.84$ for ALE and GVGAI): 
 most able techniques are those that have higher variance (as we saw in Fig.~\ref{fig:variance}). However, there is no clear correlation between ability and generality ($-0.16$ and $-0.03$ for ALE and GVGAI). This gives us the reassuring insight that the progress in these two benchmarks is not significantly due to a loss of generality.

As generality and capability could be increased (or sacrificed) independently, we can ask the question of how generality should be used for competitions, or for encouraging further progress in AI. Several options exist, such as setting a limit of generality in order to qualify for the competition, or integrate generality and ability in some compound metric. It is also important to see how generality behaves for all the participants, as if many of them are general then, because of the duality of the parameters, we will have very discriminating items. Actually, maximum generality for all agents implies maximum discrimination for all items and vice versa. Consequently, there is a risk of trying to eliminate items with low discrimination to increase the overall generality (and hence transitivity). It is important to determine whether low or negative discriminations are caused by some issues of a problem or game (e.g., it depends too much on random effects, it has strong discontinuities in the scores in terms of the effort needed to solve them, etc.), so that removing it will strengthen the evaluation, or it is because lack of generality of the population of agents, which may be solved by having more general agents in subsequent competitions. Negative discrimination can also suggest that the problem is actually an outlier, very different from the rest, and hence it may be useful to include new problems of similar characteristics to make a benchmark more general.

\section{Discussion} \label{Discussion}

The previous sections have identified four indicators to analyse the results from sets of AI problems (games) and AI systems (players). Difficulty and discrimination have been shown useful for any populational analysis of results in other areas, and they can also play an important role in AI and games. When looking at an AI problem, we see that its difficulty can be caused by several reasons (difficult underlying state representations, varying speeds and types of enemies or goals, etc.). It is however when we analyse the discrimination parameter that we at least can see whether a problem is difficult due to different reasons: \textit{(1)} it is difficult because only the good techniques are able to score well at it, or \textit{(2)} it is difficult because no technique gets it right (having a flat slope). As for ability, while usually related to average performance, it is a normalised parameter, which also takes difficulty into account. Indeed, when the discrimination of a problem is flat or negative, we cannot expect a positive monotonicity between the ability of a method and the probability of a correct response for the item. This would make many poor agents (below the difficulty of the item) getting it right and many good agents (above the item difficulty) getting it wrong.

The techniques in this paper also have some limitations. IRT needs to estimate many parameters, and it can only be applied once we have a good number of results of the respondents (controllers/algorithms) over the items (games). This is why we have chosen ALE and GVGAI to illustrate their use, as we have been able to get a relatively large results table. In the case of ALE, this can be done when the benchmark has a sufficient large number of problems and has attracted sufficient attention to get many different techniques being evaluated on it. For competitions such as GVGAI, once the participants of the competition have submitted their controllers to the game sets, the results can be used o obtain both the final rankings and the IRT parameters.

But, {\em once the parameters are estimated}, one can obtain the ability and generality for a single new agent, especially in the context of adaptive testing, without a re-estimation of all the item parameters. The obtained ability is defined on a normal scale, which is more informative and illustrative than a ranking (we can see if the winner is much better than the runner-up, for instance, in the context of the population). In general, if two editions of the same competition (or two rounds of the same competition) use the same items (or we just analysed the common items), the parameters of the first can be used to evaluate the results of the second without a re-estimation of the item parameters. Indeed, this is the recommendation when a competition is held for several rounds or editions: calculate the item parameters, and use them for the evaluation of new techniques. From time to time, the parameters can be re-evaluated when the agent population has changed significantly.

Another feature of IRT and the four parameters introduced here is that they are sample dependent. Of course we expect that things will change when we change the items (so the benchmark is actually measuring a different thing), but it is harder to understand that in single-player games the ability of an agent changes if some other new agents are included in the pool, like in adversarial games such as chess and Go (although this is not always seen as a negative thing, see e.g., \cite{balduzzi2018re}). The reason why the parameters of one agent are affected by other agents is that the notion of discrimination/difficulty in IRT is populational, so when we change the population (e.g., improving an agent), the obtained parameters will differ. Similarly, in this paper, IRT is also used to estimate the ability of a new technique, so it is also populational. As the measure of generality uses difficulty, it also becomes populational. In these conditions, the way in which agents progress can make an impact on several metrics. For instance, if agents improve on the easy items, they will become more general. The generality of good agents will help contribute making discrimination positive for more items, since discrimination is negatively affected by good agents performing poorly on easy items.

One possibility to make ability and generality less prone to changes in the population is to calibrate some parameters with the use of agents for which we have theoretical expectations, especially if we can change some of their configurations or we can tune them by some of their hyper-parameters. For instance, in \cite{prudencio2015analysis} the IRT models were derived for a random forest classifier, whose ability was gauged by the number of trees that were used in the multiclassifier.

Generality has usually been analysed from a populational point of view, starting from the very manifold in the early days of the analysis of general intelligence 
\cite{hernandez2016spearman}. 
But a more principled alternative to the sample dependence issue may be based on a non-populational notion of difficulty, using some notion of difficulty that derives from the tasks themselves \cite{hernandez2015c}. 
With this we would have a metric of generality (and capability) that would not depend on the other agents, and would not change whenever the agent population changes.

Overall, it is important to give some methodological take-aways for those general situations where we have to estimate the four parameters. First, we should wait to have a wide range of problems or games that are representative of what we want to evaluate or set as a benchmark. Adding many small variations of the same problem will affect the metrics of ability but most especially generality, as they can create clusters for which the agents can specialise. Second, we will have to wait until an important number of techniques have been applied to the benchmark, either through a literature meta-review (as we have done here for ALE) or from the results of a competition (as we have done here for GVGAI). Third, publishing the estimated parameters for problems and agents publicly is key for other researchers to use them in the evaluation of new agents or the definition of new benchmarks.

The most novel contribution in this paper is the introduction of the generality indicator, which becomes meaningful precisely because of the use of difficulty: different levels of difficulty imply different distribution of results for a method. Generality should not be understood in terms of the global variance. This paves the way for a better understanding of the G in AGI (Artificial General Intelligence), and other domains in AI that are aiming at more general-purpose AI systems. 

The four indicators, which can be obtained easily with the code we provide, can also be particularly meaningful from the viewpoint of AI benchmarks \cite{hernandezCosmos2017} and (videogame) competitions \cite{loiacono20102009,hingston2012believable,ontanon2013survey,togelius2013mario,renz2015aibirds}, as they provide a proper insight of what the games (and other tasks) are evaluating, and even whether they can be considered useless for a good evaluation in the benchmark. Also,
taking into account the long training and evaluation times of
recent computing-demanding techniques, any understanding of what the key games are (in order to reduce the size of the benchmark, specially in the hyperparameter search) can imply an important contribution for AI researchers. With the generality metric, we also have an extra parameter that can give us more information about whether the negative discriminations are caused by some pathologies of the item or because the agent population has insufficient generality.

On the other hand, we can also obtain further insight of those AI systems addressing these games (beyond their aggregated performance). It is important to determine whether the new techniques, especially those that rely on long training stages with a game, are coping well generally, and not only for a pocket of problems, but failing in some situations. This is relevant for both the AI and video game communities, but it can have broader consequences in AI \cite{martinez2018accounting,martinez2018facets} or its role of AI-based automation in the workplace \cite{fernandez2018multidisciplinary}.

Of course, other models are possible, with more or less parameters, and estimated in different ways. For instance, we want to derive more sophisticated 4-parameter IRT models using continuous inputs. In general, the specific number of parameters will depend on whether some of the variables are given or not (for instance, if we have a theoretical notion of difficulty). 
Overall, the most important insight is this dual view between AI tasks and AI systems.

\section*{Acknowledgment}

This material is based upon work supported by the U.S. Air Force Office of Scientific Research under award number FA9550-17-1-0287, the EU (FEDER), and the Spanish MINECO under grant TIN 2015-69175-C4-1-R, the Generalitat Valenciana PROMETEOII/2015/013. F. Mart\'{\i}nez-Plumed was also supported by INCIBE (Ayudas para la excelencia de los equipos de investigaci\'on avanzada en ciberseguridad), the European Commission (Joint Research Centre) HUMAINT project (Expert Contract CT-EX2018D335821-101), and Universitat Polit\`ecnica de Val\`encia (Primeros Proyectos de lnvestigaci\'on PAID-06-18). J. Hern\'andez-Orallo also received a Salvador de Madariaga grant (PRX17/00467) from the Spanish MECD for a research stay at the CFI, Cambridge, a BEST grant (BEST/2017/045) from the GVA for another research stay at the CFI, and the FLI grant RFP2-152. We want to thank David L. Dowe for his comments on an earlier version of this paper.

\bibliographystyle{apalike}
\bibliography{biblio}



\end{document}